\documentclass[runningheads]{llncs}
\usepackage{lineno,hyperref}
\usepackage[flushleft]{threeparttable}
\usepackage[noend]{algpseudocode}
\usepackage{microtype}
\usepackage{amsmath}
\usepackage{amssymb,comment}
\usepackage{mathrsfs}
\usepackage{wasysym}
\usepackage{multicol}
\usepackage{color}
\usepackage{xspace}
\usepackage{txfonts}
\usepackage{enumerate}
\usepackage{tikz}
\usepackage{nicefrac}
\usepackage{balance,multirow}
\usepackage{capt-of}
\usepackage{courier}
\usepackage{dsfont}
\usetikzlibrary{automata,arrows,positioning}
\usepackage{graphicx,marvosym}
\usepackage{xspace}
\usepackage{pifont}
\usepackage{booktabs,caption}
\usepackage{multirow}
\usepackage{graphicx}
\usepackage{subfigure}
\usepackage{makecell}
\usepackage{adjustbox}
\usepackage[algoruled,linesnumbered,noend]{algorithm2e}
\usepackage{colortbl}
\definecolor{mygray}{gray}{.85}

\newcommand{\tabincell}[2]{\begin{tabular}{@{}#1@{}}#2\end{tabular}}
\newcommand{\mb}{\mathcal{N}}
\newcommand{\ml}{\mathcal{L}}
\newcommand{\mg}{{G}}
\newcommand{\stdbool}{{\mathbb{B}}}
\newcommand{\bool}{{\mathbb{B}}_{\pm1}}
\newcommand{\fb}{{F_{\mathtt{in}}}}
\newcommand{\fo}{{F_{\mathtt{out}}}}

\newcommand{\fbb}{{f_{\mathtt{in}}}}

\newcommand{\var}{\textsf{var}}
\newcommand{\val}{\textsf{val}}
\newcommand{\hi}{\textsf{hi}}
\newcommand{\lo}{\textsf{lo}}
\newcommand{\bs}[1]{{\vec{#1}}}

\newcommand{\VAR}{\textsc{Var}}
\newcommand{\NOT}{\textsc{Not}}
\newcommand{\APPLY}{\textsc{Apply}}
\newcommand{\AND}{\textsc{And}}
\newcommand{\OR}{\textsc{Or}}
\newcommand{\XOR}{\textsc{Xor}}
\newcommand{\XNOR}{\textsc{Xnor}}
\newcommand{\EXISTS}{\textsc{Exists}}
\newcommand{\COMPOSE}{\textsc{RelProd}}

\newcommand{\ITE}{\textsc{ITE}}
\newcommand{\CONST}{\textsc{Const}}
\newcommand{\HM}{\textsf{HD}}

\newcommand{\SATALL}{\textsc{SatAll}}

\newcommand{\myrom}[1]{\uppercase\expandafter{\romannumeral#1}}

\newcommand{\fu}[1]{\textcolor{red}{#1}}
\newcommand{\yd}[1]{\textcolor{blue}{#1}}
\newcommand{\tl}[1]{\textcolor{magenta}{#1}}
\newcommand{\hide}[1]{ }

\newcommand{\tool}{{\sf BDD4BNN}\xspace}
\newcommand{\npaq}{{\sf NPAQ}\xspace}

\begin{document}
\title{\tool: A BDD-based Quantitative Analysis Framework for Binarized Neural Networks} 
%
\titlerunning{\tool}
%

\author{Yedi Zhang\inst{1} \and Zhe Zhao\inst{1} \and Guangke Chen\inst{1} \and Fu Song\inst{1} \and Taolue Chen\inst{2}}
%
%
\institute{ShanghaiTech University, Shanghai, China \and University of Surrey, Surrey, UK}

%
%
\maketitle              
\begin{abstract}
Verifying and explaining the behavior of neural networks is becoming increasingly important, especially when they are deployed in safety-critical applications. In this paper, we study verification problems for Binarized Neural Networks (BNNs), the 1-bit quantization of general real-numbered neural networks. Our approach is to encode BNNs into Binary Decision Diagrams (BDDs), which is done by exploiting the internal structure of the BNNs. In particular, we translate the input-output relation of blocks in BNNs to cardinality constraints which are then encoded by BDDs. Based on the encoding, we develop a quantitative verification framework for BNNs where precise and comprehensive analysis of BNNs can be performed. We demonstrate the application of our framework by providing quantitative robustness analysis and interpretability for BNNs. We implement a prototype tool  \tool and carry out extensive experiments which confirm the effectiveness and efficiency of our approach.
\end{abstract}
%
%

\section{Introduction}\label{sec:intro}

Deep neural networks (DNNs) have achieved human-level performance in several tasks, and are increasingly being incorporated into various application domains such as autonomous driving~\cite{Apollo} and medical diagnostics~\cite{SWS17}.
Modern DNNs usually contain a great many parameters which are typically stored as 32/64-bit floating-point numbers,  and require a massive amount of floating-point operations to compute the output for a single input~\cite{TanL19}. As a result, it is often challenging to deploy them on resource-constrained, embedded devices. To mitigate the issue, quantization, which quantizes 32/64-bit floating-points to low bit-width fixed-points (e.g., 4-bits) with little accuracy loss~\cite{GuptaAGN15}, emerges as a promising technique to reduce resource requirements. In particular, binarized neural networks (BNNs)~\cite{BNN} represent the case of 1-bit quantization using the bipolar binaries $\pm 1$. BNNs can drastically reduce memory storage and execution time with bit-wise operations,
hence substantially improve the time and  energy efficiency. BNNs have been demonstrated to achieve high accuracy for a wide variety of applications~\cite{KungZWCM18,XNOR-Net,McDanelTK17}.



DNNs have been shown to often lack robustness against adversarial samples.
Therefore, various formal techniques have been proposed to analyze DNNs,
but most of them focus on \emph{real-numbered} DNNs only.  
%
%
Verification of \emph{quantized} DNNs has not been thoroughly explored 
so far, although recent results have highlighted its importance: it was shown that a quantized DNN does not necessarily preserve the properties satisfied by the 
real-numbered DNN before quantization~\cite{DZBS19,GiacobbeHL20}.
Indeed, the fixed-point number semantics effectively yields a discrete state space for the verification of quantized DNNs whereas real-numbered DNNs feature a continuous state space. 
The discrepancy  could invalidate current verification techniques for real-numbered DNNs 
when they are directly applied to quantized counterparts (e.g., both false negative and false positive could occur).
Therefore, specialized techniques are required 
for rigorously verifying quantized DNNs.

%


Broadly speaking, the existing techniques for quantized DNNs make use of constraint solving which is based on either SAT/SMT or (reduced, ordered) binary decision diagrams (BDDs). 
A majority of work resorts to SAT/SMT solving. For the 1-bit quantization (i.e., BNNs), typically 
BNNs are transformed into Boolean formulas where SAT solving is harnessed~\cite{narodytska2018verifying,neuronFactor18,KorneevNPTBS18,Narodytska18}.
Some recent work also studies 
variants of BNNs~\cite{NarodytskaZGW20,JiaR20}, 
for instance, three-valued BNNs.
For quantized DNNs with multiple bits (i.e., fixed-points), it is natural to
encode them as quantifier-free SMT formulas, e.g., using bit-vector and fixed-point theories~\cite{BaranowskiHLNR20,GiacobbeHL20,henzinger2020scalable},
so that off-the-shelf SMT solvers can be leveraged.
%
In another direction, BDD-based approaches currently can tackle BNNs only~\cite{ddlearning19B}. 
In a nutshell, they encode a BNN and an input region as a BDD, based on which
various analyses can be performed via queries on the BDD. The crux of the approach is how to generate the BDD efficiently.
In the work \cite{ddlearning19B}, the BDD is constructed by BDD learning~\cite{Nakamura05},
thus, currently limited to toy BNNs
(e.g., 64 input size, 5 hidden neurons, and 2 output size)
with relatively small input regions.
%
%


%

On the other hand, existing work mostly focuses on \emph{qualitative} verification, which asks whether there exists an input $x$
(in a specified region) for a neural network 
such that a property  (e.g., local robustness) is violated.
In many practical applications, 
checking only the  existence is not sufficient. Indeed, for local robustness, such an (adversarial) input  almost surely exists which makes a qualitative answer less meaningful.  
Instead, \emph{quantitative} verification, which asks how often a property $\phi$ is satisfied or violated, 
is far more useful as it could provide a 
probabilistic guarantee of
the behavior of neural networks. Such a quantitative guarantee is essential to certify, for instance, certain implementations of neural network based perceptual components against 
safety standards of autonomous vehicles
~\cite{kalra2016driving,koopman2019safety}.
Quantitative analysis of  general neural networks, however, is challenging, hence received little attention and for which the results are rather limited so far.
DeepSRGR~\cite{YLLHWSXZ20} presented an abstract interpretation based quantitative robustness verification approach
for DNNs which is sound but incomplete.
For BNNs, approximate SAT model-counting
solvers ($\sharp$SAT) are leveraged~\cite{baluta2019quantitative,NarodytskaSMIM19}
based on the 
SAT encoding for the qualitative counterpart. 
Though probably approximately correct (PAC) style guarantees can be provided, 
verification cost 
is usually prohibitively high to achieve higher precision and confidence.   

\smallskip
\noindent
{\bf Main contributions.}
We propose a BDD-based framework  \tool 
to support quantitative analysis of BNNs.
The main challenge is how to efficiently
build BDDs from BNNs
~\cite{NarodytskaSMIM19}.
In contrast to previous work \cite{ddlearning19B} which is learning-based and largely treats the BNN  as a blackbox,
we 
\emph{directly} encode a BNN and the associated input region into BDDs. 
In a nutshell, a BNN is a sequential composition of multiple internal blocks and one output block.
Each block comprises a handful of layers and captures a function $f:\{+1,-1\}^{n}\rightarrow \{+1,-1\}^{m}$,
where $n$ (resp. $m$) denotes the number of inputs (resp. outputs) of the block.
Technically, the function $f$ can be alternatively rewritten as a function over the standard Boolean domain, i.e., $f:\{0,1\}^{n}\rightarrow \{0,1\}^{m}$.
A key stepping-stone of our encoding is the observation that the $i$-th output $y_i$ of the block
can be 
captured by a cardinality constraint of the form $\sum_{j=1}^n \ell_j\ge k$
such that $y_i\Leftrightarrow\sum_{j=1}^n \ell_j\ge k$,
where each literal $\ell_j$ is either $x_j$ or $\neg x_j$ for the input variable $x_j$,
and $k$ is a constant.
%
 We then present an algorithm to encode a
cardinality constraint $\sum_{j=1}^n \ell_j\ge k$
as a BDD with $O((n-k)\cdot k)$ nodes in $O((n-k)\cdot k)$ time.
As a result, the input-output relation of each block can be encoded as a BDD, the composition of which yields 
the BDD for the entire BNN. 
%
%
A distinguished advantage of our BDD encoding lies in its support of incremental encoding.  
In particular, 
when different input regions are of interest, 
there is no need to construct the BDD of the entire BNN 
from scratch. 

Encoding BNNs as BDDs enables a wide variety of applications in security analysis and decision explanation of BNNs.
In this paper, we highlight two of them within our framework, i.e.,
robustness analysis and interpretability.
It was shown that DNNs have been suffering from poor robustness to adversarial examples~\cite{SZSBEGF14,PMJFCS16,PMGJCS17}.
We consider two quantitative variants of the problem: (1) how many adversarial examples does the BNN have in the input region, and (2) how many of them are misclassified to each class?
We further provide an algorithm to incrementally compute the (locally) maximal Hamming distance within which the BNN satisfies the desired robustness properties. 

Interpretability is an issue arising as a result of the blackbox
nature of DNNs~\cite{HuangKRSSTWY20,MCB20}.
In application domains such as medical diagnosis, understanding the decisions made by DNNs has become a pressing need.
We consider two problems: (1) why some inputs are (mis)classified into a class by the BNN 
and (2) are there any essential features in the input region that are common for all samples classified into a class?

%
%
%

\smallskip
\noindent
{\bf Experimental Results.}
We implement our framework as a prototype tool \tool using the CUDD package~\cite{CUDD}, which scales to BNNs with up to 4 internal blocks, 200 hidden neurons, and 784 input size.
To the best of our knowledge, it is the first work to precisely analyze such large BNNs that go significantly beyond
the state-of-the-art.
The experimental results show that \tool is significantly more efficient and scalable than the learning-based technique~\cite{ddlearning19B}.
Furthermore, we demonstrate how \tool can be used in quantitative robustness analysis and decision explanation of BNNs.
For quantitative robustness analysis, our experimental results show that \tool is considerably ($5\times$ to $1,340\times$) faster and more accurate than the state-of-the-art
approximate $\sharp$SAT-based approach~\cite{baluta2019quantitative}. It can also  
compute precisely the distribution of predicated classes of the images in the input region as well as 
the locally maximal Hamming distances on several BNNs. For decision explanation, we show the effectiveness of \tool in computing prime-implicant explanations and essential features of the given input region for some target classes.

In general, our main contributions can be summarized as follows.
\begin{itemize}
  \item We introduce a novel algorithmic approach for encoding BNNs into BDDs that exactly preserves the semantics of BNNs,  which supports incremental encoding.
  \item We propose a framework for
  quantitative verification of BNNs and in particular, we demonstrate the robustness analysis and interpretability of BNNs.
  \item We implement the framework as an end-to-end tool \tool  
  and  conduct thorough experiments on various BNNs, demonstrating the efficiency and effectiveness of \tool.
\end{itemize}

\smallskip
\noindent
{\bf Outline.} The remainder of this paper is organized as follows.
Section~\ref{prel} briefly introduces BNNs and BDDs.
Section~\ref{sec:bnnencoding} and Section~\ref{sec:app} present our BDD-based quantitative analysis framework
and its applications respectively.
Section~\ref{sec:evaluation} reports the evaluation results.
Section~\ref{sec:relatedwork} discusses related work.
Finally, we conclude this work in Section~\ref{sec:conc}.

\begin{figure}[t]
  \centering
  \includegraphics[width=0.7\textwidth]{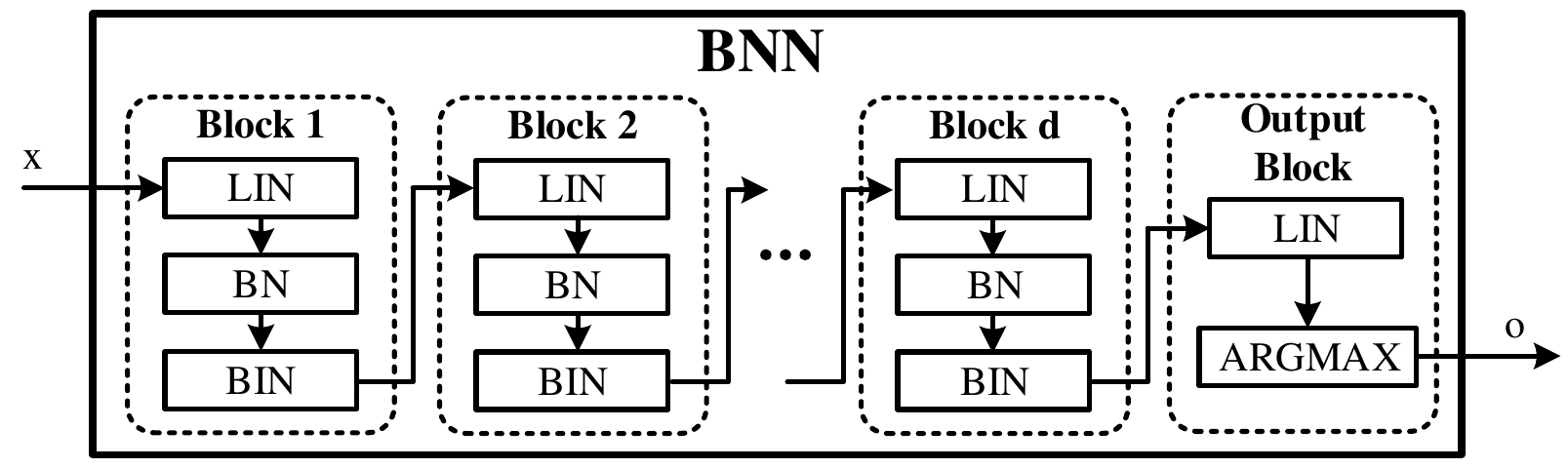}
  \caption{Architecture of a BNN with $d+1$ blocks}\label{fig:bddarch}
\end{figure}

\section{Preliminaries}\label{prel}

In this section, we briefly introduce 
binarized neural networks (BNNs) and (reduced, ordered) binary decision diagrams (BDDs).

We denote by $\mathbb{R}$, $\mathbb{N}$, $\stdbool$, and $\bool$
the set of real numbers, the set of natural numbers,
the standard Boolean domain $\{0,1\}$ and the integer set $\{+1,-1\}$.
For $n\in \mathbb{N}$, we denote by $[n]$ the set $\{1,\cdots,n\}$.
We will use $\bs{W}$, $\bs{W}', \dots$ to denote (2-dimensional) matrices,
$\bs{x}, \bs{v},\cdots$ to denote (row) vectors,
and $x,v,\dots$ to denote scalars.
We denote by $\bs{W}_{i,:}$ and $\bs{W}_{:,j}$ the $i$-th row and $j$-th  column of
the matrix $\bs{W}$. Similarly, we denote by $\bs{x}_j$ and $\bs{W}_{i,j}$
the $j$-th entry of $\bs{x}$ and $\bs{W}_{i,:}$ respectively.
In this work, Boolean values $1/0$ will be used as integers
$1/0$ in arithmetic computations without typecasting.

\subsection{Binarized Neural Networks}\label{bcnn}

A binarized neural network (BNN)~\cite{BNN} 
is a neural network where weights and activations are predominantly binarized over the domain $\bool$.
In this work, we consider feed-forward BNNs.
As shown in Figure~\ref{fig:bddarch}, a BNN can be seen as a sequential composition of several internal blocks and one output block.
Each internal block comprises
3 
layers: a linear layer (LIN), a batch normalization layer (BN), and a  binarization layer (BIN). The output block comprises a linear layer and
an  ARGMAX layer.
Note that the input/output of internal blocks and the input of the output block are all vectors over $\bool$.

\begin{definition}\label{def:bnn}
A BNN $\mb:\bool^{n_1}\rightarrow \stdbool^{s}$ with $s$ classes is given by a tuple of blocks $(t_1,\cdots,t_d,t_{d+1})$ such that \[\mb=t_{d+1}\circ t_{d}\circ\cdots \circ t_1,\]
\begin{itemize}
  \item for every $i\in[d]$, $t_i:\bool^{n_i}\rightarrow \bool^{n_{i+1}}$ is an internal block comprising
a LIN layer $t_i^{lin}$, a BN layer $t_i^{bn}$ and a BIN $t_i^{bin}$ with $t_i=t_i^{bin} \circ t_i^{bn} \circ t_i^{lin}$,

  \item $t_{d+1}:\bool^{n_{d+1}}\rightarrow \stdbool^{s}$ is the output block comprising a LIN layer $t_{d+1}^{lin}$ and an ARGMAX layer $t_{d+1}^{am}$
  with $t_{d+1}=t_{d+1}^{am} \circ t_{d+1}^{lin}$,
\end{itemize}
where $t_i^{bin}$, $t_i^{bn}$, $t_i^{lin}$ for $i\in[d]$, $t_{d+1}^{lin}$ and $t_{d+1}^{am}$ are given in Table~\ref{tab:layers}.
\end{definition}

Intuitively, a LIN layer is a linear transformation.
A BN layer following a LIN layer is used to standardize and normalize the output of the LIN layer. A BIN layer is used to binarize the real-numbered
output vector of the BN layer. In this work, we consider the sign function which is widely
used in BNNs to binarize real-numbered vectors. An ARGMAX layer follows a LIN layer and outputs the index of the  largest entry
as the predicted class which is represented by
a one-hot vector. (In case there is more than one such entry, the first one is returned.) Formally,  given a BNN $\mb=(t_1,\cdots,t_d,t_{d+1})$  and an input $\bs{x}\in \bool^{n_1}$,
$\mb(\bs{x})\in \stdbool^{s}$ is a one-hot vector in which the index of the non-zero entry is the predicated class.

\begin{table}[t]\setlength{\tabcolsep}{9pt}
	\caption{{Definitions of layers in BNNs, where $n_{d+2}=s$ and $\arg\max(\cdot)$ returns the index of the  largest entry which occurs first.}} \label{tab:layers}
	\centering
\scalebox{0.9}{  \begin{tabular}{ccccc}\toprule
     Layer & Function & Parameters & Definition \\ \midrule
     LIN & $t_i^{lin}: \bool^{n_i}\rightarrow \mathbb{R}^{n_{i+1}}$ &\tabincell{l}{Weight matrix: $\bs{W}\in \bool^{n_i\times n_{i+1}}$ \\ Bias (row) vector: $\bs{b}\in \mathbb{R}^{n_{i+1}}$} & \tabincell{c}{$t_i^{lin}(\bs{x})=\bs{y}$ where $\forall j\in [n_{i+1}]$, \\ $\bs{y}_j=\langle \bs{x}, \bs{W}_{:,j}\rangle+\bs{b}_j$ }\\  \midrule

     BN & $t_i^{bn}: \mathbb{R}^{n_{i+1}}\rightarrow \mathbb{R}^{n_{i+1}}$ &\tabincell{l}{Weight vectors: $\alpha \in \mathbb{R}^{n_{i+1}}$ \\ Bias vector: $\gamma \in \mathbb{R}^{n_{i+1}}$ \\ Mean vector:  $\mu\in \mathbb{R}^{n_{i+1}}$ \\  Std. dev. vector: $\sigma\in \mathbb{R}^{n_{i+1}}$} & \tabincell{c}{$t_i^{bn}(\bs{x})=\bs{y}$  where $\forall j\in [n_{i+1}]$, \\ $\bs{y}_j=\alpha_j\cdot(\frac{\bs{x}_j-\mu_j}{\sigma_j})+\gamma_j$}\\  \midrule

     BIN & $t_i^{bin}: \mathbb{R}^{n_{i+1}}\rightarrow \bool^{n_{i+1}}$ & - & \tabincell{c}{$t_i^{bin}(\bs{x})=\bs{y}$  where $\forall j\in [n_{i+1}]$, \\ $\bs{y}_j=\left\{
	\begin{array}{lr}
	+1, \qquad  \text{if}~~ \bs{x}_j\geq 0; \\
	-1, \qquad  \text{otherwise}.\\
	\end{array}\right.$}\\  \midrule

  ARGMAX & $t_{d+1}^{am}: \mathbb{R}^{s}\rightarrow \stdbool^{s}$ & - & \tabincell{c}{$t_{d+1}^{am}(\bs{x})=\bs{y}$  where $\forall j\in [s]$, \\ $\bs{y}_j=1 \Leftrightarrow j=\arg\max(\bs{x})$}\\ \bottomrule
   \end{tabular}}
\end{table}

\subsection{Binary Decision Diagrams}
A BDD~\cite{bryant1986graph} is a rooted acyclic directed
graph where non-terminal nodes $v$ are labeled by Boolean variables $\var(v)$ and
terminal nodes (leaves) $v$ are labeled with values $\val(v)\in\stdbool$, referred to as the 1-leaf and the 0-leaf respectively.
Each non-terminal node $v$ has two outgoing edges:
$\hi(v)$ meaning $\var(v)=1$ and  $\lo(v)$ meaning $\var(v)=0$.
We will also refer to $\hi(v)$ and $\lo(v)$ as the $\hi$ and $\lo$ children of $v$ respectively.
Moreover, assuming that $x_1,\cdots,x_m$ is the variable ordering, for each node
$v$ with $\var(v)=x_i$ and each $v'\in \{\hi(v),\lo(v)\}$ with
$\var(v')=x_j$, we have $i<j$.
In the graphical representation of BDDs, $\hi(v)$ and $\lo(v)$ are depicted by solid and dashed lines respectively.
MTBDDs are a variant of BDDs in which the terminal nodes are not restricted to be $0$ or $1$.
A BDD is  \emph{reduced}  if it (1) has only one 1-leaf and one 0-leaf,
(2) does not contain a node $v$ such that $\hi(v)=\lo(v)$,
and (3) does not contain two distinct non-terminal nodes $v$ and $v'$
such that $\var(v)=\var(v')$,  $\hi(v)=\hi(v')$ and $\lo(v)=\lo(v')$.
Hereafter, we assume that BDDs are reduced. 

Bryant~\cite{bryant1986graph} showed that BDDs can serve as a canonical form of Boolean functions.
Given a BDD over variables $x_1,\cdots,x_m$,
each non-terminal node $v$ with $\var(v)=x_i$ represents
a Boolean function $f_v=(x_i\wedge f_{\hi(v)})\vee (\neg x_i\wedge f_{\lo(v)})$.
Operations on Boolean functions can usually be efficiently implemented
via manipulating their BDD representations.
A good variable ordering is crucial for the performance of BDD manipulations while the task of finding an optimal ordering for a function is NP-hard.
To store and manipulate BDDs efficiently, nodes are stored in a 
hash table and recent computed results are stored in a cache to avoid duplicated computations.
In this work, we will use some basic BDD operations
such as $\ITE$ for If-Then-Else, $\XOR$ for exclusive-OR, $\XNOR$ for exclusive-NOR (i.e., $a~ \XNOR~ b=\neg (a~\XOR ~b)$)
and $\SATALL(f_v)$ for the set of all solutions
of the Boolean formula $f_v$. We denote by $\ml(v)$ the set $\SATALL(f_v)$.
For 
easy reference, more operations are given in Appendix~\ref{sec:bddoper}.

\section{\tool Design} \label{sec:bnnencoding}

In this section, we first present an overview of our BDD-based quantitative analysis
framework \tool, and then provide details of the key components.

\subsection{\tool Overview}
\begin{figure}[t]
  \centering
  \includegraphics[width=.9\textwidth]{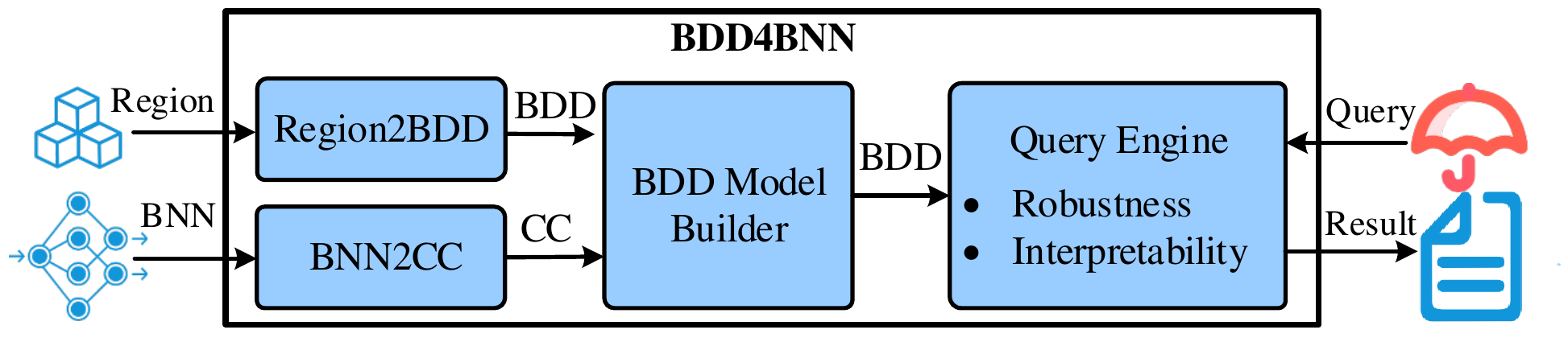}
  \caption{Overview of \tool}\label{fig:overview}
\end{figure}

An overview of \tool is depicted in Figure~\ref{fig:overview}.
\tool comprises four main components: Region2BDD, BNN2CC,
BDD Model Builder, and Query Engine. For a fixed BNN $\mb=(t_1,\cdots,t_d,t_{d+1})$ and a region $R$ of the input space of $\mb$,
\tool constructs the BDDs $(\mg_i^{out})_{i\in[s]}$ to encode the input-output relation of $\mb$ in the region $R$,
where the BDD $\mg_i^{out}$ 
corresponds to the class $i\in[s]$.
Technically, the region $R$ is partitioned into $s$ parts represented by the BDDs $(\mg_i^{out})_{i\in[s]}$.
For each query of a property, \tool analyzes $(\mg_i^{out})_{i\in[s]}$
and outputs the query result.

The general workflow of our approach is as follows.
First, Region2BDD builds up a BDD $\mg_{R}^{in}$ from the region $R$ which
represents the desired input space of $\mb$ for analysis.
Second, BNN2CC transforms each block of the BNN $\mb$ into a set of cardinality constraints (CCs) similar to~\cite{narodytska2018verifying,baluta2019quantitative}.
Third, BDD Model Builder builds the BDDs $(\mg_i^{out})_{i\in[s]}$ from all the cardinality constraints and the BDD $\mg_{R}^{in}$.
Finally, Query Engine answers  queries by analyzing the BDDs $(\mg_i^{out})_{i\in[s]}$.
Our Query Engine currently supports two types of application queries:
robustness analysis and interpretability. 

In the rest of this section, we first introduce the key sub-component CC2BDD, which provides encoding of cardinality constraints into BDDs. 
We then provide details of the components Region2BDD, BNN2CC, and BDD Model Builder.
The Query Engine will be described in Section~\ref{sec:app}.

\subsection{CC2BDD: Cardinality Constraints to BDDs}
A \emph{cardinality constraint} is a constraint of the form
$\sum_{j=1}^n \ell_j\ge k$ over a vector $\vec{x}$ of Boolean variables 
with length $n$, where the literal $\ell_j$ is either $\bs{x}_j$ or $\neg \bs{x}_j$ for each $j\in[n]$.
Note that constraints of the form $\sum_{j=1}^n \ell_j>k$, $\sum_{j=1}^n \ell_j\leq k$
and $\sum_{j=1}^n \ell_j<k$ are equivalent to $\sum_{j=1}^n \ell_j\geq k+1$, $\sum_{j=1}^n \neg \ell_j\geq n-k$
and $\sum_{j=1}^n \neg \ell_j\geq n-k+1$, respectively.
We assume that $1$ (resp. $0$) is a special cardinality constraint
that always holds (resp. never holds).

\begin{figure}[t]
	\centering
	\includegraphics[width=.8\textwidth]{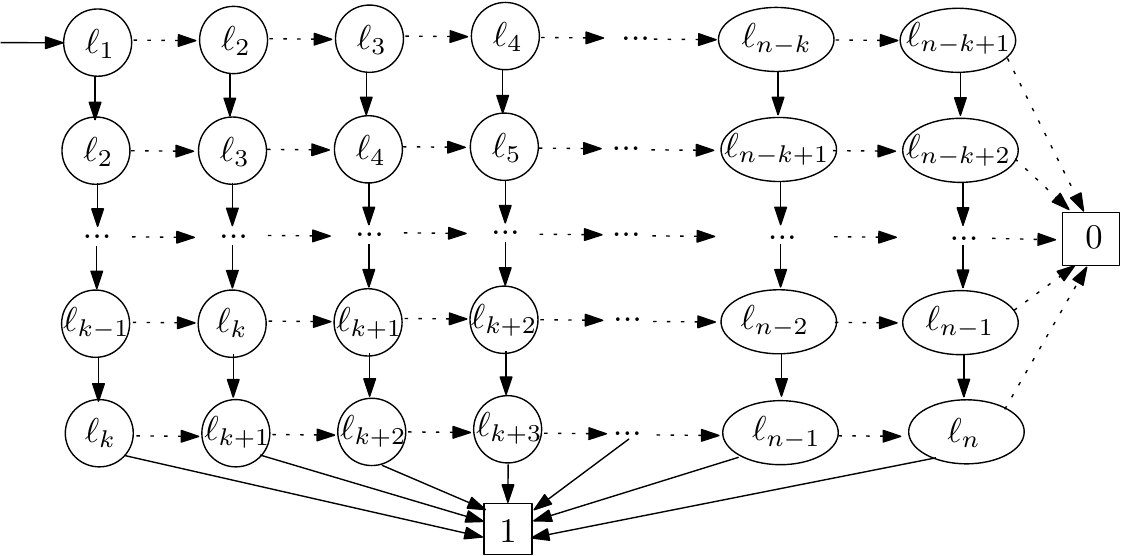}
	\caption{Graphic representation of all the solutions of $\sum_{j=1}^n \ell_j\ge k$}
	\label{fig:cc2bdddemo}
\end{figure}

To encode $\sum_{j=1}^n \ell_j\ge k$ as a BDD, we observe that all the possible solutions of
$\sum_{j=1}^n \ell_j\ge k$ can be compactly represented by a BDD-like graph shown in Figure~\ref{fig:cc2bdddemo},
where each node is labeled by a literal, and a solid (resp. dashed) edge from a node labeled by $\ell_j$ means that
the value of the literal $\ell_j$ is $1$ (resp. $0$).
Thus, each path from the $\ell_1$-node to the 1-leaf through the $\ell_j$-node (where $1\leq j\leq n$)  captures a set of valuations
where $\ell_j$ followed by a (horizontal) dashed line is set to be 0 while $\ell_j$ followed by a (vertical) solid line is set to be 1,
and all the other literals which are not along the path can take arbitrary values. Clearly, for each of these valuations, there are at least $k$ positive literals, hence the constraint  $\sum_{j=1}^n \ell_j\ge k$ holds.

%

\begin{algorithm}[t]
\small
  \SetAlgoNoLine
    \SetKwProg{myproc}{Proc}{}{}
    \myproc{{\sc CC2BDD}{$({\tt CC}: \sum_{j=1}^n \ell_j\ge k)$}}{
        $\mg_{k+1,1}=\mg_{k+1,2}=\cdots=\mg_{k+1,n-k+1}=\CONST(1)$\;
        $\mg_{1,n-k+2}=\mg_{2,n-k+2}=\cdots=\mg_{k,n-k+2}=\CONST(0)$\;
        \For{$(i=k; \ i\geq 1; \ i--)$}{
            \For{$(j=n-k+1; \  j\geq 1; \ j--)$}{
                \lIf{$(\ell_{i+j-1}==\bs{x}_{i+j-1}$)}{$\mg_{i,j}= \ITE(\bs{x}_{i+j-1},\mg_{i+1,j},\mg_{i,j+1})$}
                \lElse{$\mg_{i,j}= \ITE(\bs{x}_{i+j-1},\mg_{i,j+1},\mg_{i+1,j})$}
            }
        }
        \Return{$\mg_{1,1}$}
    }
  \caption{BDD Construction for cardinality constraints}
  \label{alg:cc2bdd}
  \end{algorithm}

Based on the above observation, we build the BDD for $\sum_{j=1}^n \ell_j\ge k$ using Algorithm~\ref{alg:cc2bdd}.
It builds a BDD for each node in Figure~\ref{fig:cc2bdddemo}, row-by-row (the index $i$ in Algorithm~\ref{alg:cc2bdd}) and
from right to left (the index $j$ in Algorithm~\ref{alg:cc2bdd}).
For each node at the $i$-th row and $j$-th column,
the label of the node must be the literal $\ell_{i+j-1}$.
We build the BDD $\mg_{i,j}= \ITE(\bs{x}_{i+j-1},\mg_{i+1,j},\mg_{i,j+1})$ if $\ell_{i+j-1}$ is of the form
$\bs{x}_{i+j-1}$ (Line 6), otherwise
we build the BDD $\mg_{i,j}= \ITE(\bs{x}_{i+j-1},\mg_{i,j+1},\mg_{i+1,j})$ (Line 7).
Finally, we obtain the BDD $\mg_{1,1}$ that encodes the solutions of
$\sum_{j=1}^n \ell_j\ge k$.

\begin{lemma}\label{lem:cc2bdd}
For each cardinality constraint $\sum_{j=1}^n \ell_j\ge k$,
a BDD $\mg$ with $O((n-k)\cdot k)$  nodes can be computed in $O((n-k)\cdot k)$ time
such that $\ml(\mg)$ is the set of all the solutions of $\sum_{j=1}^n \ell_j\ge k$.
\end{lemma}

\subsection{Region2BDD: Input Regions to BDDs}\label{sec:region2bdd}

In this paper, we consider the following two 
types of input regions.
\begin{itemize}
  \item \emph{Input region based on Hamming distance}. For an input $\bs{u}\in \bool^{n_1}$ and an integer $r\geq 0$, $R(\bs{u},r)$ denotes the set $\{\bs{x}\in \bool^{n_1}\mid \HM(\bs{x},\bs{u})\leq r\}$, where
   $\HM(\bs{x},\bs{u})$ denotes the Hamming distance between $\bs{x}$ and $\bs{u}$. Intuitively, $R(\bs{u},r)$ includes the input vectors which differ from $\vec{u}$ by at most $r$ positions. 

  \item \emph{Input region with fixed indices}. For an input $\bs{u}\in \bool^{n_1}$ and a set of indices $I\subseteq [n_1]$, $R(\bs{u},I)$ denotes the set  $\{\bs{x}\in \bool^{n_1}\mid \forall i\in [n_1]\setminus I.\ \bs{u}_i=\bs{x}_i\}$. Intuitively, $R(\bs{u},I)$ includes the input vectors which differ from $\vec{u}$  only at the indices from $I$. 
\end{itemize}
Note that both $R(\bs{u},n_1)$ and $R(\bs{u},[n_1])$ denote the entire input space $\bool^{n_1}$.

Recall that each input sample is an element from $\bool^{n_1}$. 
To represent the region $R$ by a BDD,
we  transform each value $\pm1$ into a Boolean value $1/0$.
To this end, for each input $\bs{u}\in \bool^{n_1}$,
we create a new sample $\bs{u}^{(b)}\in \stdbool^{n_1}$ such that
for every $i\in [n_1]$, $\bs{u}^{(b)}_i=2\bs{u}_i-1$.
Therefore, $R(\bs{u},r)$ and $R(\bs{u},I)$ will be represented by
$R(\bs{u}^{(b)},r)$ and $R(\bs{u}^{(b)},I)$, respectively.
The transformation functions $t_i^{lin}$, $t_i^{bn}$, $t_i^{bin}$ and $t_{d+1}^{am}$ of the LIN, BN, BIN, and ARGMAX layers (cf. Table~\ref{tab:layers})
will be handled accordingly. Note that for convenience,
vectors over the Boolean domain $\stdbool$ may be directly given by $\bs{u}$ or $\bs{x}$ when it is clear from the context.

\smallskip
\noindent
{\bf Region Encoding under Hamming distance}. Given an input $\bs{u}\in\stdbool^{n_1}$ and an integer $r$,
the region $R(\bs{u},r)$ can be expressed by a cardinality constraint
$\sum_{j=1}^{n_1} \ell_j\leq r$ (which is equivalent to
$\sum_{j=1}^{n_1} \neg\ell_j\geq n_1-r$), where for every $j\in[n_1]$, $\ell_j=\bs{x}_j$ if $\bs{u}_j=0$, otherwise  $\ell_j=\neg \bs{x}_j$.
For instance, consider $\bs{u}=(1,1,1,0,0)$ and $r=2$, we have:
\[\HM(\bs{u},\bs{x})=1\oplus \bs{x}_1+1\oplus \bs{x}_2+1\oplus \bs{x}_3+ 0\oplus \bs{x}_4+0\oplus \bs{x}_5=
\neg \bs{x}_1+\neg \bs{x}_2+\neg \bs{x}_3+  \bs{x}_4+ \bs{x}_5.\]
Thus, $R((1,1,1,0,0),2)$ can be expressed by the cardinality constraint
$\neg \bs{x}_1+\neg \bs{x}_2+\neg \bs{x}_3+  \bs{x}_4+ \bs{x}_5\leq 2$,
or equivalently $ \bs{x}_1+ \bs{x}_2+ \bs{x}_3+  \neg\bs{x}_4+ \neg\bs{x}_5\geq 3$.

By Algorithm~\ref{alg:cc2bdd},  the cardinality constraint of $R(\bs{u},r)$ can be encoded by
the BDD $G^{in}_{\bs{u},r}$,
such that $\ml(G^{in}_{\bs{u},r})=R(\bs{u},r)$.
Following Lemma~\ref{lem:cc2bdd}, we get that:

\begin{lemma}
For every input region $R$ given by an input $\bs{u}\in\stdbool^{n_1}$ and an integer $r$,
a BDD $G^{in}_{\bs{u},r}$ with $O(r\cdot (n_1-r))$ nodes can be computed in $O(r\cdot(n_1-r))$ time
such that $\ml(G^{in}_{\bs{u},r})=R(\bs{u},r)$.
\end{lemma}

\noindent
{\bf Region Encoding under fixed indices}.
Given an input $\bs{u}\in\stdbool^{n_1}$ and a set of indices $I\subseteq [n_1]$,
the region $R(\bs{u},I)=\{\bs{x}\in \stdbool^{n_1}\mid \forall i\in [n_1]\setminus I.\ \bs{u}_i=\bs{x}_i\}$
can be represented by the following BDD
\[G^{in}_{\bs{u},I}\triangleq\AND_{i\in [n_1]\setminus I} \Big( (\bs{u}_i==1)? \VAR(\bs{x}_i): \NOT(\VAR(\bs{x}_i))\Big).\]

\begin{lemma}
For every input region $R$ given by an input $\bs{u}\in\stdbool^{n_1}$ and a set of indices $I\subseteq [n_1]$,
a BDD $G^{in}_{\bs{u},I}$ with $O(n_1-|I|)$ nodes can be computed in $O(n_1)$ time
such that $\ml(G^{in}_{\bs{u},I})=R(\bs{u},I)$.
\end{lemma}

\subsection{BNN2CC: BNNs to Cardinality Constraints}\label{sec:bnn2cc}
As mentioned before, to encode the BNN $\mb=(t_1,\cdots,t_d,t_{d+1})$ as BDDs,
we transform the BNN $\mb$ into cardinality constraints from which the desired BDDs $(\mg_i^{out})_{i\in[s]}$ are constructed.
To  this end, we first transform each internal block
$t_i:\bool^{n_i}\rightarrow \bool^{n_{i+1}}$ into $n_{i+1}$ cardinality constraints, each of which
corresponds to one of the outputs of $t_i$.
Then we  transform the output block $t_{d+1}:\bool^{n_{d+1}}\rightarrow \stdbool^{s}$
into $s(s-1)$ cardinality constraints, where
one output class yields $(s-1)$ cardinality constraints.

For each vector-valued function $t$, 
we denote by $t_{\downarrow j}$ the (scalar-valued) function
returning the $j$-th entry of the output of $t$.

\smallskip
\noindent
{\bf Transformation for internal blocks.}
Consider the internal block $t_i:\bool^{n_i}\rightarrow \bool^{n_{i+1}}$ for $i\in[d]$.
Recall that for every $j\in[n_{i+1}]$ and $\bs{x} \in \bool^{n_i}$,
$t_{i\downarrow j}(\bs{x})  = t_i^{bin}( t_i^{bn}(\langle \bs{x}, \bs{W}_{:,j}\rangle+\bs{b}_j))$,
and each value $\pm1$ of an input $\bs{u}\in\bool^{n_1}$ is replaced by $1/0$ (cf. Section~\ref{sec:region2bdd}).
%
To be consistent, the function $t_{i\downarrow j}:\bool^{n_i}\rightarrow \bool$ is reformulated as the function $t_{i\downarrow j}^{(b)}:\stdbool^{n_i}\rightarrow \stdbool$
such that for every $\bs{x}\in \stdbool^{n_i}$,
$t_{i\downarrow j}^{(b)}(\bs{x})  = t_i^{bin}( t_i^{bn}(\langle 2\bs{x}-\bs{1}, \bs{W}_{:,j}\rangle+\bs{b}_j)),$
where $\bs{1}$ denotes the vector of $1$'s with the width $n_i$.

Let $C_{i,j}$ be the following cardinality constraint:
\[C_{i,j}\triangleq \left\{
\begin{array}{lr}
  \sum_{k=1}^{n_i}\ell_k \geq \lceil \frac{1}{2}\cdot (n_i+\mu_j-\bs{b}_j-\frac{\gamma_j\cdot \sigma_j}{\alpha_j}) \rceil, & \text{if} \ \alpha_j>0;\\
  1, & \text{if} \ \alpha_j=0\wedge \gamma_j\geq 0;\\
  0, & \text{if} \ \alpha_j=0\wedge \gamma_j< 0;\\
  \sum_{k=1}^{n_i}\neg\ell_k \geq \lceil \frac{1}{2}\cdot (n_i-\mu_j+\bs{b}_j+\frac{\gamma_j\cdot \sigma_j}{\alpha_j}) \rceil,   & \text{if} \ \alpha_j<0;
\end{array}\right.\]
where for every $k\in [n_i]$, $\ell_k$ is $\bs{x}_k$ if $\bs{W}_{k,j}=+1$, and $\ell_k$ is $\neg\bs{x}_k$ if $\bs{W}_{k,j}=-1$.

\begin{proposition}\label{prop:inblk2cc}
$t_{i\downarrow j}^{(b)}\Leftrightarrow C_{i,j}$.
\end{proposition}

\smallskip
\noindent
{\bf Transformation for the output block.}
For the output block $t_{d+1}:\bool^{n_{d+1}}\rightarrow \stdbool^{s}$,
since $t_{d+1}=t_{d+1}^{am} \circ t_{d+1}^{lin}$, then for every $j\in[s]$,
we can reformulate $t_{d+1\downarrow j}:\bool^{n_{d+1}}\rightarrow \stdbool$ as the function $t_{d+1\downarrow j}^{(b)}:\stdbool^{n_{d+1}}\rightarrow \stdbool$
such that for every $\bs{x}\in \stdbool^{n_{d+1}}$,
$t_{d+1\downarrow j}^{(b)}(\bs{x})=t_{d+1\downarrow j}(2\bs{x}-\bs{1})$.

For every $j'\in [s]\setminus\{j\}$, we define the cardinality constraint $C_{d+1,j'}$
as follows:
\[C_{d+1,j'}\triangleq \left\{
\begin{array}{l}
\sum_{k=1}^{n_{d+1}} \ell_{d+1,k} \geq \frac{1}{4}(\bs{b}_j'-\bs{b}_j+ \sum_{k=1}^{n_{d+1}} (\bs{W}_{k,j}-\bs{W}_{k,j'}))+1+\sharp{\tt Neg}, ~~~~~~~~~~~~~~~~~~~~~~~   ~~~~~~~~~~~~~~~~~  \\
\hfill  ~~~~~~~~~~~~~~ \text{if} \ j'<j \text{ and } \frac{1}{4}(\bs{b}_j'-\bs{b}_j+ \sum_{k=1}^{n_{d+1}} (\bs{W}_{k,j}-\bs{W}_{k,j'})) \ \text{is an integer};\\ \\
\sum_{k=1}^{n_{d+1}} \ell_{d+1,k} \geq \lceil\frac{1}{4}(\bs{b}_j'-\bs{b}_j+ \sum_{k=1}^{n_{d+1}} (\bs{W}_{k,j}-\bs{W}_{k,j'}))\rceil+\sharp {\tt Neg},  \hfill \text{otherwise}; \\
\end{array}\right.\]
where $\sharp {\tt Neg}=|\{k\in [n_{d+1}] \mid \bs{W}_{k,j}-\bs{W}_{k,j'}=-2\}|$, $\ell_{d+1,k}$ is $\bs{x}_{d+1,k}$ if $\bs{W}_{k,j}-\bs{W}_{k,j'}=+2$,
$\ell_{d+1,k}$ is  $\neg \bs{x}_{d+1,k}$ if $\bs{W}_{k,j}-\bs{W}_{k,j'}=-2$,
and $\ell_{d+1,k}$ is $0$ if $\bs{W}_{k,j}-\bs{W}_{k,j'}=0$.

\begin{proposition}\label{prop:outblk2cc}
$t_{d+1\downarrow j}^{(b)} \Leftrightarrow  \bigwedge_{j'\in [s], j'\neq j} C_{d+1,j'}$. 
\end{proposition}


For each internal block $t_i:\bool^{n_i}\rightarrow \bool^{n_{i+1}}$,
we denote by BNN2CC$(t_i)$ the cardinality constraints $\{C_{i,1},\cdots, C_{i,n_{i+1}}\}$.
For each output class $j\in[s]$,
we denote by BNN2CC$^j(t_{d+1})$ the cardinality constraints
$\{C_{d+1,1},\cdots C_{d+1,j-1},C_{d+1,j+1},\cdots, C_{d+1,s}\}$.
By applying the above transformation to all the blocks of the BNN $\mb=(t_1,\cdots,t_d,t_{d+1})$,
we obtain its cardinality constraint form $\mb^{(b)}=(t_1^{(b)},\cdots,t_d^{(b)},t_{d+1}^{(b)})$
such that for each $i\in [d]$, $t_i^{(b)}=\text{BNN2CC}(t_i)$,
and $t_{d+1}^{(b)}=(\text{BNN2CC}^1(t_{d+1}),\cdots, \text{BNN2CC}^s(t_{d+1}))$.
Given an input $\bs{u}\in\stdbool^{n_1}$, we denote by $\mb^{(b)}(\bs{u})$ the index
$j\in[s]$ such that all the cardinality
constraints in $\text{BNN2CC}^j(t_{d+1})$ hold under the valuation $\bs{u}$.
It is straightforward to verify:

\begin{theorem}
$\bs{u}\in\bool^{n_1}$ is classified into the class $j$ by the BNN $\mb$ iff $\mb^{(b)}(\bs{u}^{(b)})=j$.
\end{theorem}

\subsection{BDD Model Builder}
The construction of the BDDs $(\mg_i^{out})_{i\in [s]}$
from the BNN $\mb^{(b)}$ and the input region $R$
%
is done iteratively throughout the blocks. 
Initially, the BDD for the first block is built, which can be seen as the input-output relation for the first internal block. In the $i$-th iteration, as the input-output relation of the first $(i-1)$ internal blocks has been encoded into the BDD, we compose this BDD with the BDD for the block $t_i$ which is built from its cardinality constraints $t_i^{(b)}$, resulting in the BDD for the first $i$ internal blocks.
Finally, we obtain the BDDs $(\mg_i^{out})_{i\in [s]}$ of the BNN $\mb$, with respect to the input region $R$.

\smallskip
\noindent
{\bf Design choice.} There are several design choices for efficiency consideration which we discuss as follows.
First of all, to encode the input-output relation of an internal block $t_i$ into BDD from its cardinality constraints $t_i^{(b)}=\{C_{i,1},\cdots, C_{i,n_{i+1}}\}$,
it amounts to compute $\AND_{j\in[n_{i+1}]}\text{CC2BDD}(C_{i,j})$.
A simple and straightforward approach is to initially compute a BDD $G=\text{CC2BDD}(C_{i,1})$
and then iteratively compute the conjunction $G=\AND(G, \text{CC2BDD}(C_{i,j}))$
of $G$ and $\text{CC2BDD}(C_{i,j})$ for $2\leq j\leq n_{i+1}$.

Alternatively, we use a divide-and-conquer strategy to recursively compute the BDDs for the first half and the second half of the cardinality constraints respectively, and then apply the AND-operation.  
%
Our preliminary experimental results show that the latter approach often performs better (about 2 times faster) than the former one, although they generate the same BDD.

Second,  constructing the BDD directly from the cardinality constraints $t_i^{(b)}=\{C_{i,1},\cdots, C_{i,n_{i+1}}\}$
becomes prohibitively  costly when $n_i$ and $n_{i+1}$ are large, as the BDDs $\text{CC2BDD}(C_{i,j})$
for $j\in[n_{i+1}]$ need to consider all the inputs in $\stdbool^{n_i}$.
To improve efficiency, we apply feasible input propagation. Namely, when we construct the BDD for the block $t_{i+1}$, we only consider its possible inputs with respect to the output of the block $t_i$.
Our preliminary experimental results show that 
the optimization could significantly improve the efficiency of  the BDD construction.


Third, instead of encoding the input-output relation of the BNN $\mb$ as a sole BDD or MTBDD,
we opt to use a family of $s$ BDDs $(\mg_i^{out})_{i\in [s]}$, each of which corresponds to one output class of $\mb$.
Recall that each output class $i\in[s]$ is represented by $(s-1)$ cardinality constraints.
Then, we can build a BDD  $\mg_i$ for the output class $i$, similar to the BDD construction for internal blocks.
By composing $\mg_i$ with the BDD of the entire internal blocks, we obtain the BDD $\mg_i^{out}$.
Building a single BDD or MTBDD for the BNN is possible from $(\mg_i^{out})_{i\in [s]}$, but our approach gives the flexibility
especially when a specific  target class is interested, which is common for robustness analysis.

\begin{algorithm}[t]
\small
\SetAlgoNoLine
\SetKwProg{myproc}{Proc}{}{}
\myproc{{\sc BNN2BDD}{$({\tt BNN}: \mb=(t_1,\cdots,t_d,t_{d+1}),\ {\tt Region}: \ R(\bs{u},\tau))$}}{
     $\mg^{in}=\mg^{in}_{\bs{u},\tau}$ (cf. Section~\ref{sec:region2bdd});
     $\mb^{(b)}=(t_1^{(b)},\cdots,t_d^{(b)},t_{d+1}^{(b)})$ (cf. Section~\ref{sec:bnn2cc})\;
     \For{$(i=1; \ i\leq d; \ i++)$}{
         $\mg'=${\sc Block2BDD}$(t_i^{(b)},\mg^{in},i)$\;
         $\mg^{in}=\EXISTS(\mg',\bs{x}^i)$ \tcp*{$\bs{x}^i$ denote input variables of $t_i^{(b)}$}
         $\mg=(i==1) \ ? \ \mg' : \COMPOSE(\mg,\mg')$\;
     }
     \For{$(i=1; \ i\leq s; \ i++)$}{
         $\mg_i=${\sc Block2BDD}$(t_{d+1\downarrow i}^{(b)},\mg^{in},d+1)$\;
         $\mg_i^{out}=\COMPOSE(\mg_i,\mg)$\;
     }
     \Return{$(\mg_i^{out})_{i\in[s]}$}
}

\smallskip
\myproc{{\sc Block2BDD}{$({\tt CCs}:\{C_m,\cdots,C_n\},{\tt InputSpace}:\mg^{in}, {\tt BlkIndex}: i)$}}{
    \If{$n==m$}{
        $\mg_1=$CC2BDD$(C_m)$ (cf. Algorithm~\ref{alg:cc2bdd})\;
        $\mg=\AND(\mg_1,\mg^{in})$\;
        \lIf{$i\neq d+1$}{$\mg=\XNOR(\bs{x}_m^{i+1},\mg)$}
    }
    \Else{
        $\mg_1=${\sc Block2BDD}$(\{C_m,\cdots,C_{\lfloor\frac{n-m}{2}\rfloor+m}\},\mg^{in},i)$\;
        $\mg_2=${\sc Block2BDD}$(\{C_{\lfloor\frac{n-m}{2}\rfloor+m+1},\cdots,C_n\},\mg^{in},i)$\;
        $\mg=\AND(\mg_1,\mg_2)$\;
    }
    \Return{$\mg$}
}
  \caption{BDD Construction for BNNs}
  \label{alg:dnn2bdd}
  \end{algorithm}


\smallskip
\noindent
{\bf Overall algorithm.}
The overall BDD construction procedure is shown in~Algorithm~\ref{alg:dnn2bdd}.
Given a BNN $\mb=(t_1,\cdots,t_d,t_{d+1})$ with $s$ output classes and an input region $R(\bs{u},\tau)$, the algorithm outputs the BDDs $(\mg_i^{out})_{i\in[s]}$,
encoding the input-output relation of the BNN $\mb$
with respect to the input region $R(\bs{u},\tau)$.

The procedure {\sc BNN2BDD} first builds the BDD representation $\mg^{in}_{\bs{u},\tau}$ of the input region $R(\bs{u},\tau)$
and the cardinality constraints from BNN $\mb^{(b)}$ (Line 1).
The first for-loop builds a BDD encoding the input-output relation of the entire internal blocks w.r.t. $\mg^{in}_{\bs{u},\tau}$.
The second for-loop builds the BDDs $(\mg_i^{out})_{i\in[s]}$, each of which
encodes the input-output relation of the entire BNN
for a class $i\in[s]$ w.r.t. $\mg^{in}_{\bs{u},\tau}$.
The procedure {\sc Block2BDD} receives the cardinality
constraints $\{C_m,\cdots,C_n\}$, a BDD $\mg^{in}$ representing the  feasible inputs of the block
and the block index $i$ as inputs, and returns a BDD $\mg$.
If $i=d+1$, namely, the cardinality
constraints $\{C_m,\cdots,C_n\}$ are from the output block,
the resulting BDD $\mg$ encodes the subset of $\mg^{in}_{\bs{u},\tau}$ that satisfy all the cardinality
constraints $\{C_m,\cdots,C_n\}$.
If $i\neq d+1$, then
the BDD $\mg$ encodes the input-output relation of the Boolean function $f_{m,n}$ such that
for every $\bs{x}^{i}\in \ml(G^{in})$, $f_{m,n}(\bs{x}^{i})$
is the truth vector of the cardinality
constraints $\{C_m,\cdots,C_n\}$ under the valuation $\bs{x}^{i}$.
When $m=1$ and $n=n_{i+1}$, $f_{m,n}$ is the same as $t_i^{(b)}$, hence
$\ml(\mg)=\{\bs{x}^i\times \bs{x}^{i+1}\in \mg^{in}\times \stdbool^{n_{i+1}}\mid t_i^{(d)}(\bs{x}^i)=\bs{x}^{i+1}\}$.
Detailed explanation refers to Appendix~\ref{sec:explanation}.

\begin{theorem}
Given a BNN $\mb$ with $s$ output classes and an input region $R(\bs{u},\tau)$,
we can compute $s$ BDDs $(\mg_i^{out})_{i\in [s]}$ such that
the BNN $\mb$ classifies an input $\bs{x}\in R(\bs{u},\tau)$ into the class $i\in[s]$ iff $\bs{x}^{(b)}\in\ml(G_i^{out})$.
\end{theorem}

Algorithm~\ref{alg:dnn2bdd} 
explicitly involves $O(d+s)$ $\COMPOSE$-operations,
$O(s^2+\sum_{i\in[d]} n_i)$ $\AND$-operations and $O(d)$ $\EXISTS$-operations.

\section{Applications: Robustness Analysis and Interpretability}\label{sec:app}
In this section, we 
present two applications within \tool, i.e., robustness analysis and interpretability of BNNs.

\subsection{Robustness Analysis}
\begin{definition}
Given a BNN $\mb$ and an input region $R(\bs{u},\tau)$,
the BNN is (locally) robust w.r.t. the region $R(\bs{u},\tau)$ if each sample $\bs{x}\in R(\bs{u},\tau)$
is classified into the same class as the ground-truth class of $\bs{u}$.

An adversarial example in the region $R(\bs{u},\tau)$ is a sample $\bs{x}\in R(\bs{u},\tau)$ such that
$\bs{x}$ is classified into a class, that differs from the ground-truth class of $\bs{u}$.
\end{definition}

As mentioned in Section~\ref{sec:intro}, qualitative verification which checks whether a BNN is robust or not
is insufficient in many practical applications.
In this paper, we are interested in  \emph{quantitative} verification of robustness which asks \emph{how many adversarial examples are there
in the input region of the BNN for each class}.  To answer this question, given a BNN $\mb$ and an input region $R(\bs{u},\tau)$,
we first obtain the BDDs $(\mg_i^{out})_{i\in[s]}$ by applying Algorithm~\ref{alg:dnn2bdd}
and then count the number of adversarial examples for each class in the input region $R(\bs{u},\tau)$.
Note that counting adversarial examples amounts to computing $|R(\bs{u},\tau)|- |\ml(\mg_g^{out})|$,
where $g$ denotes the ground-truth class of  $\bs{u}$, and $|\ml(\mg_g^{out})|$ can be computed in time 
$O(|\mg_{g}^{out}|)$.

In some applications, more refined analysis is needed. 
For instance, 
it may be acceptable
to 
misclassify a dog as a cat, but unacceptable to 
misclassify a tree as
a car. This suggests that the robustness of BNNs may depend on the classes
to which samples are misclassified. To capture this,
we consider  the notion of targeted robustness.

\begin{definition}
Given a BNN $\mb$, an input region $R(\bs{u},\tau)$, and the class $t$,
the BNN is \emph{$t$-target-robust} w.r.t. the region $R(\bs{u},\tau)$ if every sample $\bs{x}\in R(\bs{u},\tau)$
is never classified into the class $t$. (Note that we assume that the ground-truth class of $\bs{u}$ differs from the class $t$.)
\end{definition}

The quantitative verification problem of $t$-target-robustness of a BNN asks
\emph{how many adversarial examples in the input region $R(\bs{u},\tau)$ are misclassified to the class $t$ by the
BNN $\mb$}.
To answer this question, we first obtain the BDD $\mg_t^{out}$ by applying Algorithm~\ref{alg:dnn2bdd}
and then count the number of adversarial examples by computing 
$|\ml(\mg_t^{out})|$.

Note that, if one wants to compute the (locally) maximal safe Hamming distance that satisfies a robustness property for an input sample (e.g., the proportion of adversarial examples is below a threshold), 
our framework can incrementally compute such a distance 
without constructing the BDD models of the entire BNN from scratch.

\begin{definition}
Given a BNN $\mb$, input region $R(\bs{u},r)$ and threshold $\epsilon\geq 0$,
$r_1$ is the (locally) maximal safe Hamming distance of $R(\bs{u},\tau)$, if one of the follows holds:
\begin{itemize}
	\item if $Pr(R(\bs{u},r))>\epsilon$, then $Pr(R(\bs{u},r_1))\leq\epsilon$
	and $Pr(R(\bs{u},r'))>\epsilon$ for $r':r_1< r'<r$;
	\item if $Pr(R(\bs{u},r))\leq \epsilon$, then $Pr(R(\bs{u},r_1+1))>\epsilon$ and $Pr(R(\bs{u},r'))\leq\epsilon$ for $r':r< r'\leq r_1$;
\end{itemize}
where $Pr(R(\bs{u},r))$ is the probability $\frac{\sum_{i\in[s].i\neq g}|\ml(\mg^{out}_i)|}{|R(\bs{u},r)|}$
for $g$ being the ground-truth class of $\bs{u}$.
\end{definition}

Algorithm~\ref{alg:maxrobust} shows the procedure to incrementally compute the maximal safe Hamming distance
for a given threshold $\epsilon\geq 0$, input region $R(\bs{u},r)$, and ground-truth class $g$ of $\bs{u}$.
Remark that $Pr(R(\bs{u},r))$ may not be 
monotonic w.r.t. the Hamming distance $r$.

\begin{algorithm}[t]
\small
\SetAlgoNoLine
\SetKwProg{myproc}{Proc}{}{}
\myproc{{\sc MaxHD}{$({\tt BNN}: \mb=(t_1,\cdots,t_d,t_{d+1}),\ {\tt Region}: \ R(\bs{u},r),  {\tt Threshold}: \epsilon, {\tt Class}: g)$}}{
     $(\mg_i^{out})_{i\in[s]}=${\sc BNN2BDD}$(\mb,R(\bs{u},r))$\;
     \If(\tcp*[h]{decrease $r$}){$(\frac{\sum_{i\in[s].i\neq g}|\ml(\mg^{out}_i)|}{|R(\bs{u},r)|}>\epsilon)$}{
        \While{$(r\geq 0)$}  {
            $r=r-1$\;
            $(\mg_i^{out})_{i\in[s]}=(\AND(G^{in}_{\bs{u},r},\mg_i^{out}))_{i\in[s]}$\;
            \lIf{$(\frac{\sum_{i\in[s].i\neq g}|\ml(\mg^{out}_i)|}{|R(\bs{u},r)|}\leq \epsilon)$}{\Return{$r$}}
        }
     }
     \Else(\tcp*[h]{increase $r$}){
        \While(\tcp*[h]{$n_1$ is the input size of the BNN $\mb$} ){$(r\leq n_1)$}
        {
            $r=r+1$\;
            $(B_i^{out})_{i\in[s]}=${\sc BNN2BDD}$(\mb,R(\bs{u},r)\setminus R(\bs{u},r-1))$\;
            $(\mg_i^{out})_{i\in[s]}=(\OR(B_i^{out},\mg_i^{out}))_{i\in[s]}$\;
            \lIf{$(\frac{\sum_{i\in[s]}|\ml(\mg^{out}_i.i\neq g)|}{|R(\bs{u},r)|}>\epsilon)$}{\Return{$r-1$}}
        }
     }
     \Return{$r$}
}
  \caption{Compute the maximal safe Hamming distance}
  \label{alg:maxrobust}
  \end{algorithm}

\subsection{Interpretability}
%
In general, interpretability addresses the question of \emph{why some inputs in the
input region are 
(mis)classified by the BNN into a specific class?}
We consider the interpretability of BNNs using two complementary explanations, i.e.,
prime implicant explanations and essential features.

\begin{definition}
Given a BNN $\mb$, an input region $R(\bs{u},\tau)$ and a class $g$,
a \emph{prime implicant explanation} (PI-explanation) of decisions made by the BNN $\mb$ on the inputs $\ml(\mg_g^{out})$ 
is a minimal set of literals $\{\ell_1,\cdots,\ell_k\}$ such that
for every $\bs{x}\in R(\bs{u},\tau)$, if $\bs{x}$ satisfies $\ell_1\wedge\cdots \wedge\ell_k$, then
$\bs{x}$ is classified into the class $g$  by the BNN $\mb$.
\end{definition}

Intuitively, a PI-explanation $\{\ell_1,\cdots,\ell_k\}$
indicates that $\{\var(\ell_1),\cdots,\var(\ell_k)\}$ are key features, namely, if fixed, the predication is guaranteed no matter how the remaining features change.
Remark that there may be more than one PI-explanation for a set of inputs $\ml(\mg_g^{out})$. 
When 
$g$ is set to be the class of the benign input $\bs{u}$,
a PI-explanation on $\mg_g^{out}$ 
suggests why
these samples are classified into $g$ by the BNN $\mb$.


\begin{definition}
Given a BNN $\mb$, an input region $R(\bs{u},\tau)$ and a class $g$, the
\emph{essential features} for the inputs $\ml(\mg_g^{out})$ 
are literals $\{\ell_1,\cdots,\ell_k\}$ such that
every $\bs{x}\in R(\bs{u},\tau)$, if $\bs{x}$ is classified into the class $g$ 
by the BNN $\mb$,
then $\bs{x}$ satisfies $\ell_1\wedge\cdots \wedge\ell_k$.
\end{definition}

Intuitively, the essential features $\{\ell_1,\cdots,\ell_k\}$ denote the
key features such that all samples $\bs{x}\in R(\bs{u},\tau)$ that are classified into the class $g$ by the BNN $\mb$
must agree on these features. Essential features differ from PI-explanations,
where the former can be seen as a necessary condition, while the latter can be seen as
a sufficient condition.

BDD libraries (e.g., CUDD~\cite{CUDD}) usually provide APIs
to identify prime implicants  (e.g., Cudd\_bddPrintCover and Cudd\_FirstPrime)
and essential variables (e.g., Cudd\_FindEssential).
Therefore, prime implicants and essential features can be computed via queries on the BDDs $(\mg_i^{out})_{i\in[s]}$.

\section{Evaluation}\label{sec:evaluation}
We have implemented our framework as a prototype tool \tool based on  the CUDD package~\cite{CUDD}.
\tool is implemented with Python as the front-end to pre-process BNNs and C++ as the back-end to perform the BDD encoding and analysis.
In this section, we report the experimental results, including
BDD encoding, robustness analysis, and interpretability.
Because of space restriction, the results of BDD encoding and robustness analysis
of BNNs with fixed indices are given in Appendix~\ref{sec:expFI}.

\smallskip
\noindent
\textbf{Experimental Setup.} The experiments were conducted on a machine with Intel Xeon Gold 5118 2.3GHz CPU, 64-bit Ubuntu 20.04 LTS operating systems, 128G RAM.
Each BDD encoding executed on one core limited by 8-hour.

\smallskip
\noindent
\textbf{Benchmarks.} We use the PyTorch (v1.0.1.post2) deep learning platform provided by NPAQ~\cite{baluta2019quantitative}
to train and test BNNs. We trained 12 BNN models (P1-P12) with varying sizes using the MNIST dataset~\cite{MNIST}.
The MNIST dateset contains 70,000 gray-scale 28 $\times$ 28 images (60,000 for training and 10,000 for testing) of
handwritten digits with 10 classes. In our experiments, we downscale the images ($28\times 28$) to some selected input size $n_1$ (i.e., the corresponding image is of the size $\sqrt{n_1}\times\sqrt{n_1}$) and then binarize the normalized pixels of the images.

Details of the BNN models are listed in Table~\ref{tab:bench}, each of which has 10 classes (i.e., $s=10$).
Column 1 shows the name of the BNN model.
Column 2 shows the architecture of the BNN model,
where $n_1:\cdots:n_{d+1}:s$ denotes that
the BNN model has $d+1$ blocks, $n_1$ inputs
and $s$ outputs; the $i$-th block for $i\in[d+1]$ has
$n_i$ inputs and $n_{i+1}$ outputs with $n_{d+2}=s$.
Recall that each internal block has 3 layers while the output block has 2 layers.
Therefore, the number of layers ranges from 5  to 14,
the dimension of inputs ranges from 9 to 784, and
the number of hidden neurons per linear layer ranges from 10 to 100.
Column 3 shows the accuracy of the BNN model on the test set of the  MNIST dataset.
(We can observe that the accuracy increases with 
the size of inputs, the number of layers, and the number of hidden neurons per layer.)
We randomly choose 10 images (shown in Figure~\ref{fig:784inputs} in Appendix) from the test set of the MNIST dataset (one image per class)
to evaluate our approach.


\begin{table}[t]
	\centering
	\caption{BNN benchmarks}
	\label{tab:bench}\setlength{\tabcolsep}{9pt} 
	\scalebox{0.85}{\begin{tabular}{ccc|ccc}
		\toprule
		Name & Architecture & Accuracy & Name &  Architecture & Accuracy  \\ \midrule
		\rowcolor{gray!20}
		P1 & 9:20:10 & 12.23\%&  P7 & 100:100:10 & 75.16\% \\
		
		P2 & 16:32:10 & 28.63\%& P8 & 100:50:20:10 & 71.1\%\\
		\rowcolor{gray!20}
		P3 & 16:64:32:10 & 25.14\% & P9 & 100:100:50:10 & 77.37\%\\
		
		P4 & 36:15:10:10 &27.12\% & P10 & 100:50:30:30:10 & 80.63\%\\
		\rowcolor{gray!20}
		P5 & 64:10:10 & 49.16\% &  P11 & 784:30:50:50:50:10  & 88.23\% \\
		
		P6 & 100:50:10 & 73.25\% &  P12 & 784:50:50:50:50:10  & 86.95\% \\
		\bottomrule
	\end{tabular}}\vspace*{-3mm}
\end{table}

\subsection{Performance of BDD Encoding}
We evaluate \tool on the BNNs listed in Table~\ref{tab:bench} using different input regions.

\begin{table}[t]
	\centering\setlength{\tabcolsep}{9pt}
	\caption{BDD encoding using full input space}
	\label{tab:uniEnc}
	\scalebox{0.8}{\begin{tabular}{crrrrc}
		\toprule
		Name & P1 & P2 & P3 & P4 & P5\\ \midrule
		Time (s) & $\approx$0 & 0.78 & 28.21 & 10924.51 & Timeout\\
		$|\mg|$ &288  & 18,864 & 17,636  & 152,830,875 & - \\
		\bottomrule
	\end{tabular}}  \vspace*{-4mm}
\end{table}

\begin{table}[t]
	\centering\setlength{\tabcolsep}{4pt}
	\caption{BDD encoding under Hamming distance}
	\label{tab:locEncHD}
	\scalebox{0.75}{\begin{tabular}{c|rr|rr|rr|rr|rr}
			\toprule
			 & \multicolumn{2}{c|}{r=2} &\multicolumn{2}{c|}{r=3} &\multicolumn{2}{c|}{r=4} &\multicolumn{2}{c|}{r=5} &\multicolumn{2}{c}{r=6} \\
			~& Time(s) &$| \mg |$&Time(s)&$| \mg |$&Time(s)&$| \mg |$&Time(s)&$| \mg |$&Time(s)&$| \mg |$\\
			\midrule \rowcolor{gray!20}
			P5 & 0.01 & 1,559 & 0.03 & 9,795 & 0.11 & 36,796  & 0.74  & 176,107 & 2.94 & 592,104  \\ 
			
			P6 & 0.25 & 4,670 & 4.17 & 84,037 & 109.26 &1,018,571 & 2,292.5 & 11,375,842 & (5) 17,811 & 41,883,970  \\ 
			\rowcolor{gray!20}
			P7 & 0.65 & 5,295 & 22.70 & 106,754 & 652.78 & 1,575,722 &(1) 17,399  & 16,163,078 & [10] & -\\  
			
			P8 & 0.14 & 6,147 & 1.95 & 125,226 & 44.51 & 1,668,027 & 1,146.8 & 20,519,582 & (1) 12,491 & 172,369,297 \\ 
			\rowcolor{gray!20}
			P9 & 1.99 & 6,139 & 63.30 & 136,126 & 1,428.6 & 2,005,666 &[1](3) 17,039  & 29,323,244 & [10] & - \\ 
			
			P10 & 0.30  & 4,630 & 4.87 & 100,054 & 101.41 & 1,603,920 & 1,909.9 & 19,844,299  & (5) 20,484 & 173,316,483\\ 
			\rowcolor{gray!20}
			P11 & 5.52 & 3,128 & 5.73 & 22,120 & 6.60 & 86,413 & 11.63 & 556,774 & 238.2 & 2,881,468\\
			
			P12 & 12.4 & 5,693 & 12.87 & 49,996 & 16.92 & 493,820 & 403.09 & 5,739,602 & (1) 11,058 & 16,241,733\\
		\bottomrule
	\end{tabular}}  \vspace*{-4mm}
\end{table}

\smallskip
\noindent
{\bf BDD encoding using full input space.}
We evaluate \tool on the BNNs (P1--P5), where $\bool^{n_1}$ is used as the input region.
The results are shown in Table~\ref{tab:uniEnc}, where $|\mg|$ denotes
the number of BDD nodes in the BDD manager. 
We can observe that both the execution time and the number of BDD nodes increase
with 
the size of BNNs. 

\smallskip
\noindent
{\bf BDD encoding under Hamming distance.}
We evaluate \tool on the relatively large BNNs (P5--P12). In this case,
an input region is given by one of the 10 images and a Hamming distance $r$ ranging from 2 to 6.
The average results are shown in Table~\ref{tab:locEncHD}, where $[i]$ (resp. $(i)$) indicates
the number of cases that \tool runs out of memory (resp. time).
Overall, the execution time and the number of BDD nodes increase
with $r$. \tool succeeded on all the cases when $r\leq 4$,
75 cases out of 80 when $r=5$, and 48 cases out of 80 when $r=6$.
We observe that the execution time and number of BDD nodes increase with 
the number of hidden neurons (P6 vs. P7, P8 vs. P9, and P11 vs. P12),
while the effect of the number of layers is diverse (P6 vs. P8 vs. P10, and P7 vs. P9).
From P9 and P10, we observe that the number of hidden neurons per layer is likely the key impact factor
of the efficiency of \tool. Interestingly, our tool \tool works well on BNNs with large input sizes (i.e., on P11 and P12).

These results demonstrate the efficiency and scalability of \tool on BDD encoding of BNNs.
We remark that, compared with the learning-based approach~\cite{ddlearning19B},
our approach is considerably more efficient and scalable.
For instance, the learning-based approach takes 403 seconds to encode a BNN with 64 input size, 5 hidden neurons, and 2 output size
when $r=6$, while ours takes about 3 seconds even for a larger network P5.



\subsection{Robustness Analysis}
We evaluate \tool on the robustness of BNNs, including
robustness analysis under different input regions and maximal safe Hamming distance computing.



\smallskip
\noindent
\textbf{Robustness verification with Hamming distance.}
We evaluate \tool on BNNs (P7, P8, P9, and P11) using the 10 images. The input regions are given by the Hamming distance $r$ ranging from 2 to 4, resulting
in $120$ instances.
To the best of our knowledge,
\npaq~\cite{baluta2019quantitative} is the only work that supports quantitative robustness verification
of BNNs to which we compare \tool.  Recall that \npaq only provides  PAC-style guarantees. Namely, it sets a tolerable error $\varepsilon$ and a confidence parameter $\delta$. The final estimated results of \npaq have the bounded error $\varepsilon$
with confidence of at least $1-\delta$, i.e.,
\begin{equation}\label{equ:errorrate}
  Pr[(1+\varepsilon)^{-1} {\tt RealNum}\leq {\tt EstimatedNum}\leq (1+\varepsilon){\tt RealNum}]\geq 1-\delta
\end{equation}
In our experiments, we set $\varepsilon=0.8$ and $\delta=0.2$, as done in~\cite{baluta2019quantitative}.

\begin{table}[t]
	\centering\setlength{\tabcolsep}{6pt}
	\caption{Robustness verification under Hamming distance}\label{tab:robustHam}
	\begin{adjustbox}{width=1\textwidth,center}
	\begin{tabular}{  c  c | rrr | rrr | rr }
		\toprule
		  & \multirow{2}*{{r}}  & \multicolumn{3}{c|}{\npaq~\cite{baluta2019quantitative}} & \multicolumn{3}{c|}{\tool} & \multicolumn{2}{c}{ Diff } \\
		 &  & \#(Adv) &  Time(s) &Pr(adv) & \#(Adv)  &  Time(s) & Pr(adv) & \#(Adv) & Speed Up\\
		\midrule
		& 2 & 875  & 271.07 & 17.32\% & 1,806 & 0.65 & 35.76\% & 106.4\%& 416\\ \rowcolor{gray!20}
		\cellcolor{white}P7 & 3 & 39,587& 919.88  & 23.74\% &65,054  &22.71 & 39.01\% & 64.33\%& 40\\
		& 4 & 1,023,798 & 3,862.0 & 25.04\% &1,501,691 & 661.79 &  36.73\% & 46.68\%& 5\\
		\midrule
		& 2 & 1,601 &  187.78 &31.70\% & 2,261  & 0.14 & 44.76\%& 41.22\% &1,340\\ \rowcolor{gray!20}
	
		\cellcolor{white}P8 & 3 & 66,562 & 396.45 & 39.92\% & 64,372 & 1.96 & 38.60\%  & -3.29\%&201\\
	
		& 4 &1,636,070& 1,861.7 & 40.02\% & 1,829,103 & 45.0 & 44.74\% & 11.80\%&40\\
		\midrule
		 & 2 &1,214 & 363.44 & 24.03\% & 1,406 & 1.99 & 27.84\% & 15.82\%& 182\\\rowcolor{gray!20}
		
		\cellcolor{white}P9& 3 &51,464 & 3,763.6 & 30.86\% & 42,901 & 63.31 & 25.73\%  &-16.64\%&58\\
		
		& 4 & 1,316,181 & (1) 9,007.8 & 32.20\% & 3,968,609 & 1,505.0 & 97.08\% &  201.5\%  & 5 \\
		\midrule
		& 2 & 12,083 & 3,831.0 & 3.93\% & 28,736 & 5.52 & 9.34\% & 137.8\% & 693\\\rowcolor{gray!20}
		
		\cellcolor{white} P11 & 3 & 0 &(2) 4,634.2  & 0\% & 0 &5.68  &0\% &  - & 815 \\
	
		& 4 & 0 &(2) 7,979.1  &0\% &  0 &6.38 &0\%  & - & 1,250 \\
		\bottomrule
	\end{tabular}
	\end{adjustbox}  \vspace*{-4mm}
\end{table}

\begin{figure}[t]
	\centering
	\subfigure[P8 under Hamming distance with $r=2$]{\label{fig:p8hdr2}
		\begin{minipage}[b]{0.45\textwidth}
			\includegraphics[width=.92\textwidth]{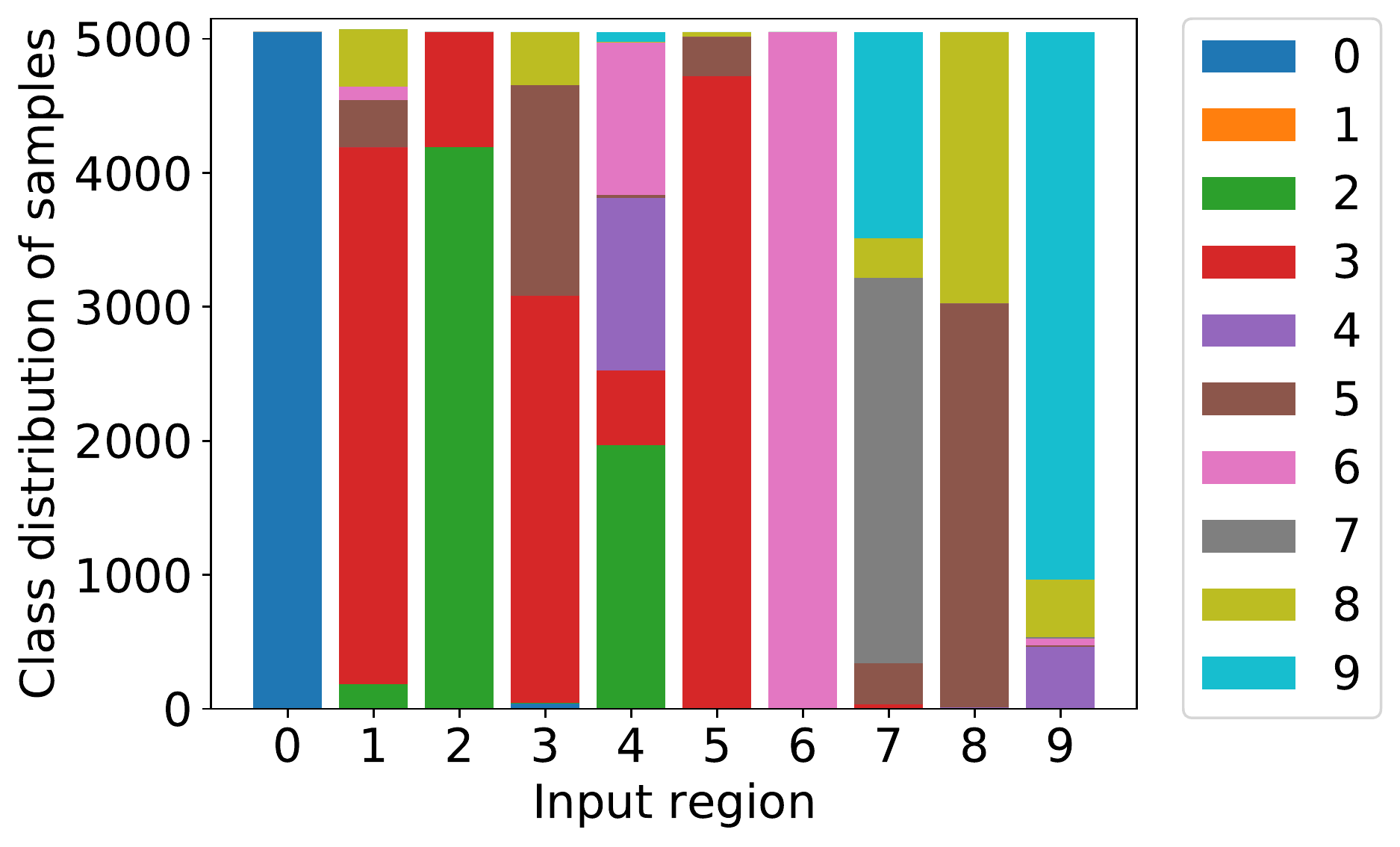}
		\end{minipage}	
	}\vspace{-3mm}
\subfigure[P8 under Hamming distance with $r=3$]{\label{fig:p8hdr3}
		\begin{minipage}[b]{0.45\textwidth}
			\includegraphics[width=.92\textwidth]{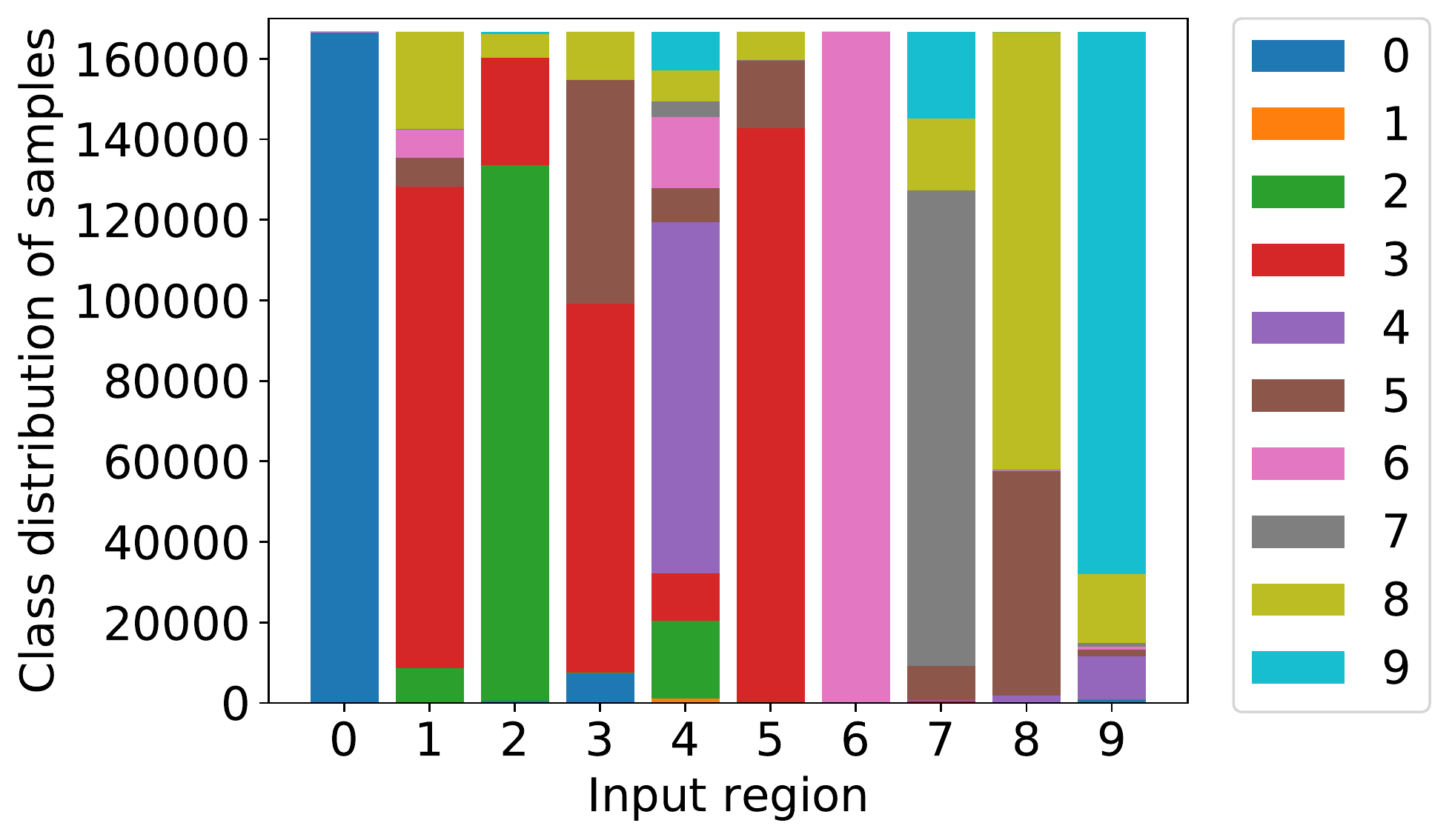}
		\end{minipage}	
	}
\subfigure[P11 under Hamming distance with $r=2$]{\label{fig:p11hdr2}
	\begin{minipage}[b]{0.45\textwidth}
			\includegraphics[width=.92\textwidth]{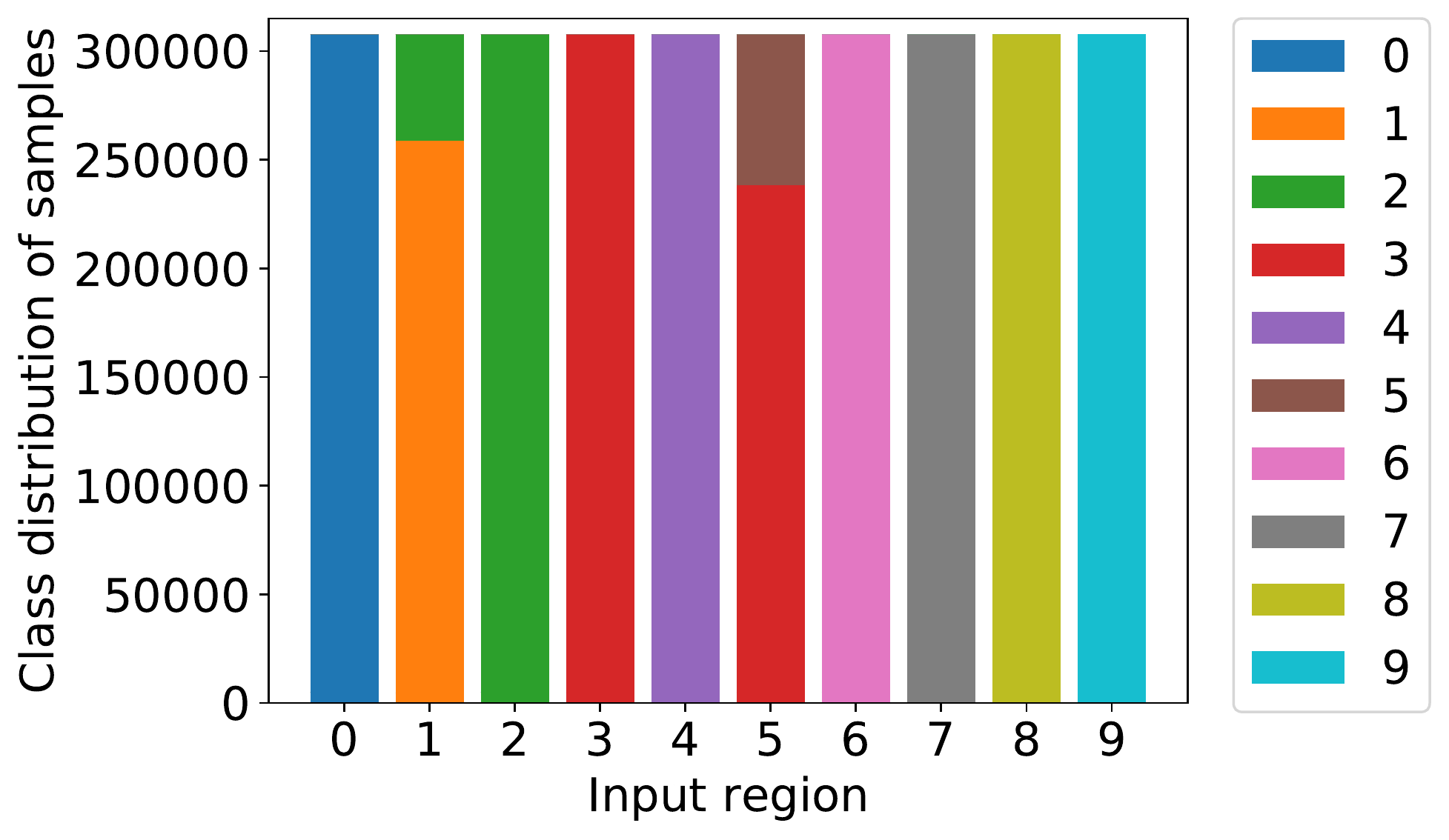}
		\end{minipage}	
	}	
\subfigure[Distribution of error rates of \npaq]{\label{fig:risingfallingrate}
		\begin{minipage}[b]{0.45\textwidth}
			\includegraphics[width=.92\linewidth]{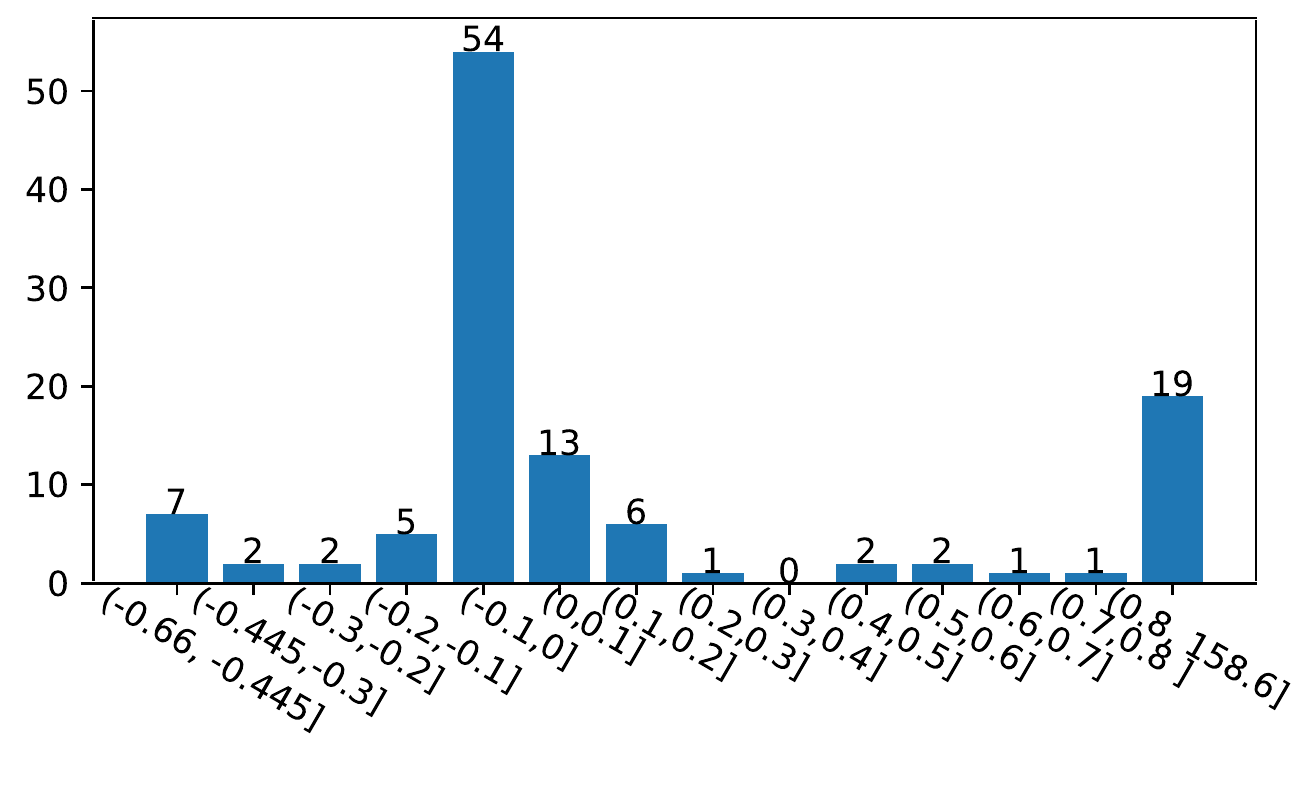}
		\end{minipage}	
	}  \vspace*{-3mm}
	\caption{Details of robustness verification with Hamming distance}\vspace*{-4mm}
\end{figure}
The results on the average of the images are shown in Table~\ref{tab:robustHam}.
\npaq ran out of time on 5 instances (which occur in P9 with $r=4$
and P11 with $r=3$ and $r=4$), while \tool successfully verified all the 120 instances.
Table~\ref{tab:robustHam} only shows the results of 115 instances that can be solved by \npaq.
Columns 3, 4, and 5 (resp. 6, 7, and 8) show the number of adversarial examples,
the execution time, and the proportion of adversarial examples in the input region.
Column 9 shows the error rate $\frac{{\tt RealNum}-{\tt EstimatedNum}}{{\tt EstimatedNum}}$,
where ${\tt RealNum}$ is from our result, and ${\tt EstimatedNum}$ is from \npaq.
Column 10 shows the speedup of  \tool compared with \npaq.
Remark that the numbers of adversarial examples are $0$
for P11 on input regions with $r=3$ and $r=4$ that can be solved by \npaq.
There do exist input regions for P11 that cannot be solved by \npaq but
have adversarial examples (see below).
On 
BNNs that were solved by both
\npaq and \tool, \tool is significantly ($5\times$ to $1,340\times$) faster 
and more accurate than \npaq. From Table~\ref{tab:locEncHD} and Table~\ref{tab:robustHam}, we also found that
most of the verification time is spent on BDD encoding while
the rest is usually less than 10 seconds.

\smallskip
\noindent
\textbf{Details of robustness and targeted robustness.}  Figure~\ref{fig:p8hdr2} (resp. Figure~\ref{fig:p8hdr3} and Figure~\ref{fig:p11hdr2}) depicts the distributions of classes on
P8 with Hamming distance $r=2$ (resp. P8 with $r=3$ and P11 with $r=2$), where on the x-axis 
$i=0,\cdots,9$ denotes
the input region that is within the respective Hamming distance to the 
image of digit $i$ (called  $i$-region).
We can observe that P8 is robust for the $0$-region when $r=2$\footnote{Note P8 is not robust for $0$-region when $r=3$, which is hard to be visualized in Figure~\ref{fig:p8hdr3} due to the small number of adversarial examples.} and robust for the $6$-region
when $r=2$ and $r=3$, but is not robust for the other regions.
Most of the adversarial examples in the $1$-region and $5$-region
are misclassified into the digit $3$ by P8.
P11 is not robust for the $1$-region or  the $5$-region, but
is robust for all the other regions.
Though P8 and P11 are not robust on some input regions,  indeed they
are $t$-target-robust for many target classes $t$, e.g.,
P11 is $t$-target-robust for the $1$-region when $t\neq 2$,
and the $5$-region when  $t\neq 3$.
(The raw data are given in  Tables~\ref{tab:P8robustHD}
and \ref{tab:P11robustHD} in Appendix.)

\smallskip
\noindent
\textbf{Quality validation of \npaq.} Figure~\ref{fig:risingfallingrate} shows the distribution of error rates of \npaq,
where the x-axis is the range of the error rate
and the y-axis is the corresponding number of instances.
There are 19 instances where the estimated number of adversarial examples exceeds $(1+\epsilon)$ of the real number of the adversarial examples and 7 instances where the estimated number of adversarial examples is less than $(1+\epsilon)^{-1}$ of the real number of the adversarial examples. This means that out of 115 instances, only in 89 instances the estimated number is within the allowed range, which is less than $1-\delta=0.8$.

\smallskip
\noindent
\textbf{Maximal safe Hamming distance.}
As a representative of such an analysis, we evaluate \tool on 4 BNNs (P7, P8, P9, and P11) with 10 images for 2 robustness thresholds ($\epsilon=0$ and $\epsilon=0.03$).
The initial Hamming distance $r$ is $3$.
Intuitively, $\epsilon=0$ (resp. $\epsilon=0.03$) means that up to 0\% (resp. 3\%) samples in the input region can be adversarial.

 Table~\ref{tab:SD} shows the  results, where
columns SD and  Time give the maximal safe Hamming distance and the execution time, respectively.
\tool solved 74 out of 80 instances. (For the remaining 6 instances,
\tool ran out of time or memory, but it was still able to compute a larger safe Hamming distance.)
We can observe that the maximal safe Hamming distance increases with the threshold $\epsilon$ on several
BNNs and input regions.
We can also observe that P11 is more robust than others, which is consistent with their accuracies (cf. Table~\ref{tab:bench}).
Remark that $\text{SD}= -1$ indicates that the input image 
itself is misclassified. 
\subsection{Interpretability}
To demonstrate the ability of \tool on interpretability, we consider the analysis of
the BNN P12 and the image $\bs{u}$ of digit 1.

\begin{figure}[t]
	\centering
	\subfigure[EFs for class 2]{\label{fig:Ess1to2}
		\begin{minipage}[b]{0.23\textwidth}
			\includegraphics[width=1\textwidth]{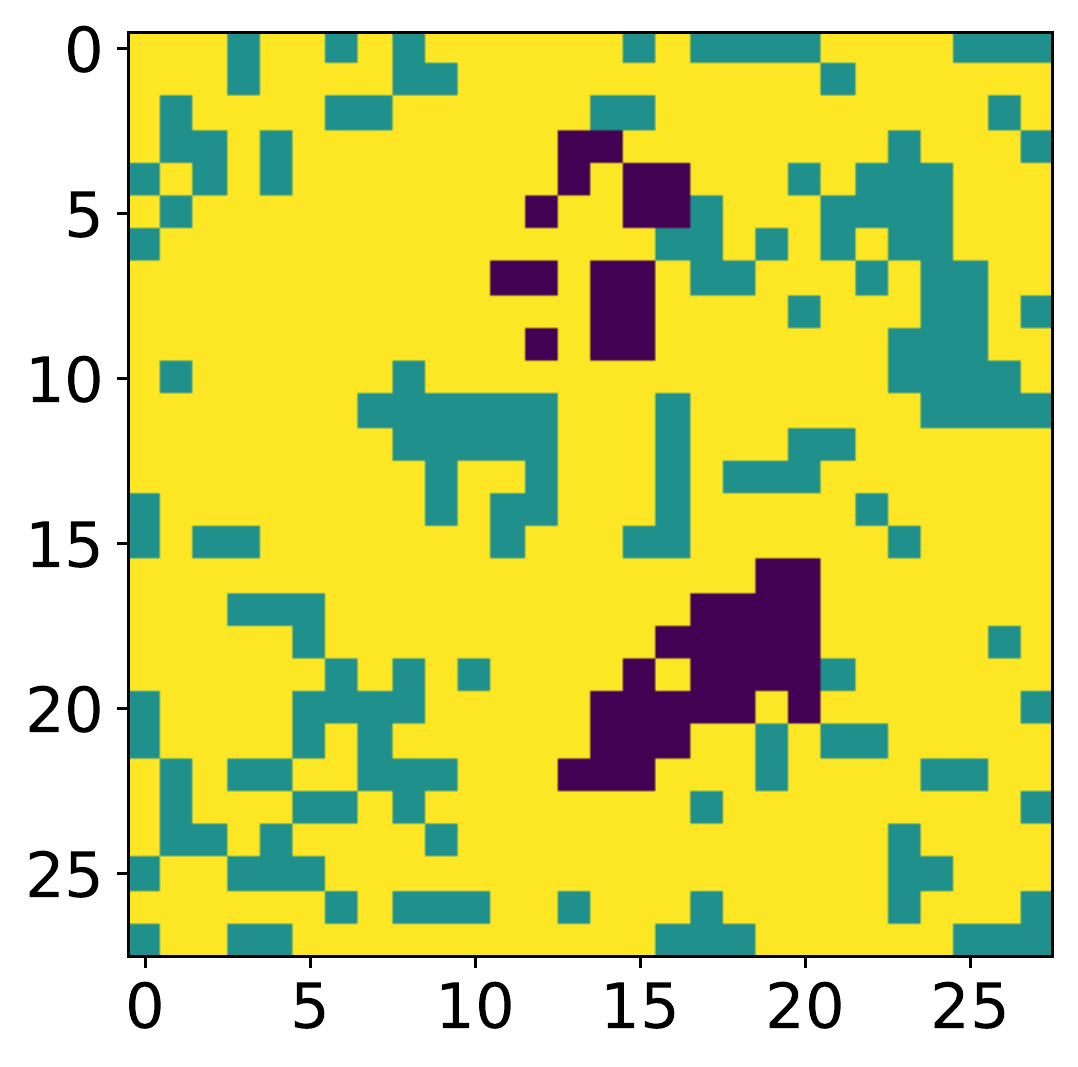}
		\end{minipage}	
	}
	\subfigure[EFs for class 5]{\label{fig:Ess1to5}
		\begin{minipage}[b]{0.23\textwidth}
			\includegraphics[width=1\textwidth]{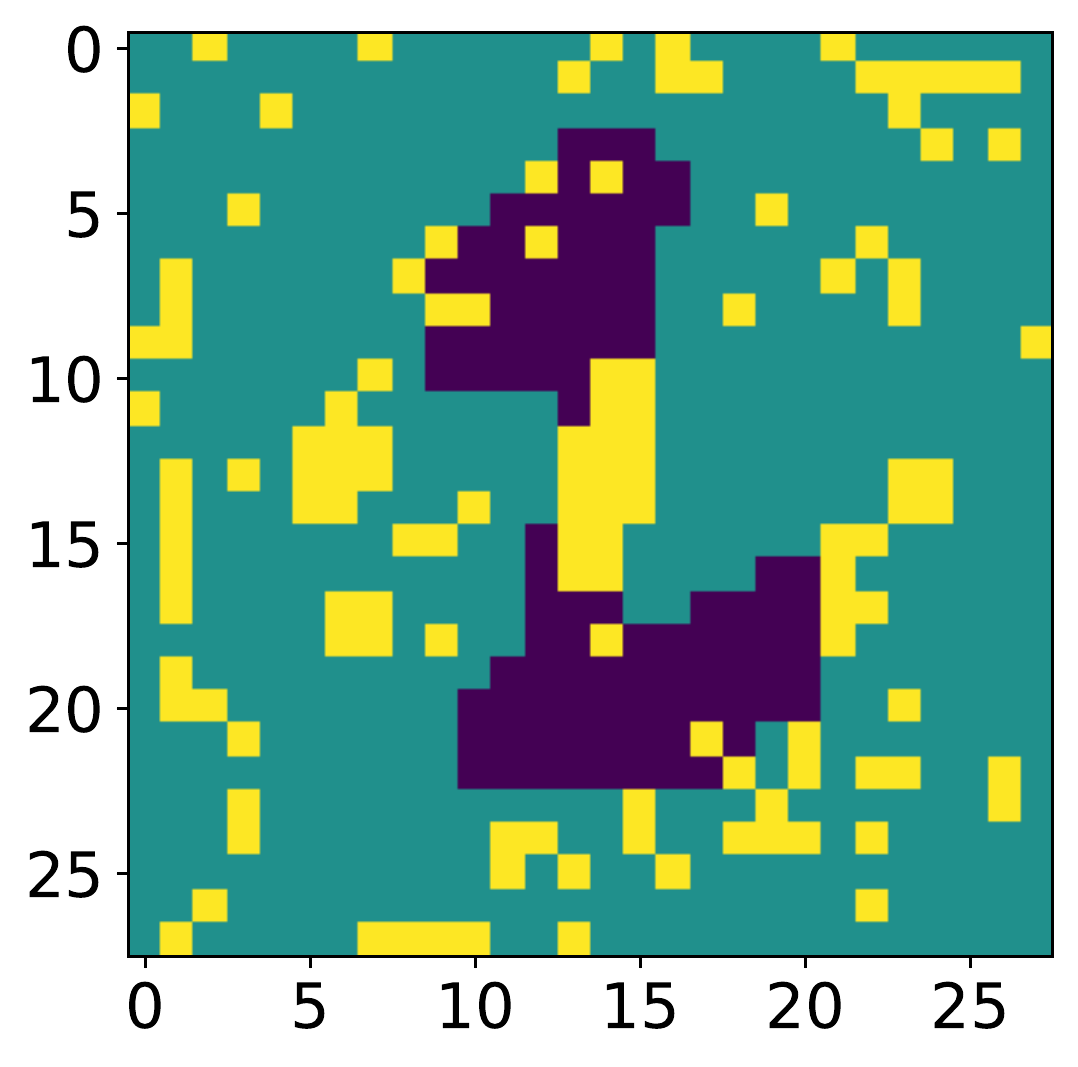}
		\end{minipage}	
	}
	\subfigure[PI for class 2]{\label{fig:PI1to2}
		\begin{minipage}[b]{0.23\textwidth}
			\includegraphics[width=1\linewidth]{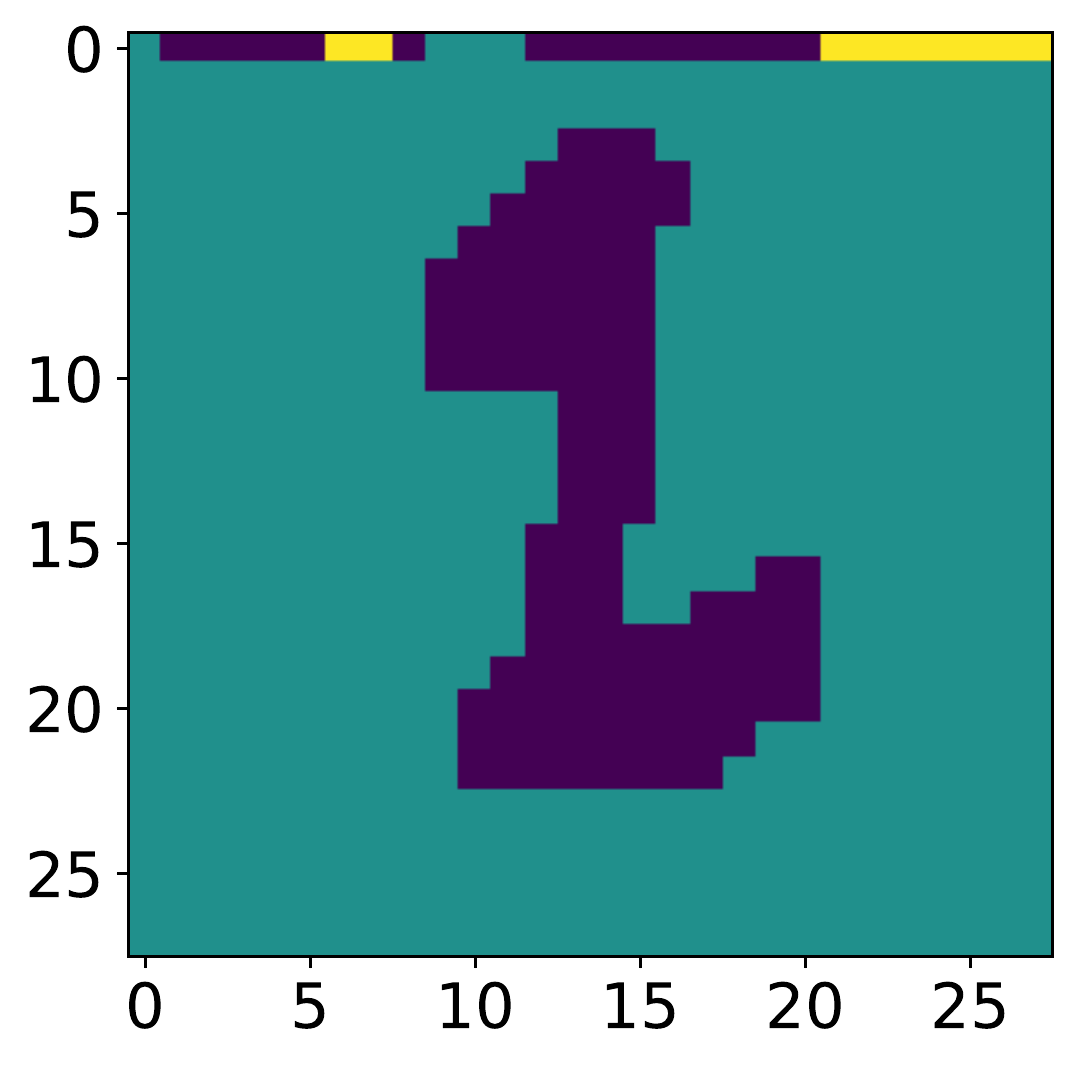}
		\end{minipage}	
	}
	\subfigure[PI for class 5]{\label{fig:PI1to5}
		\begin{minipage}[b]{0.23\textwidth}
			\includegraphics[width=1\textwidth]{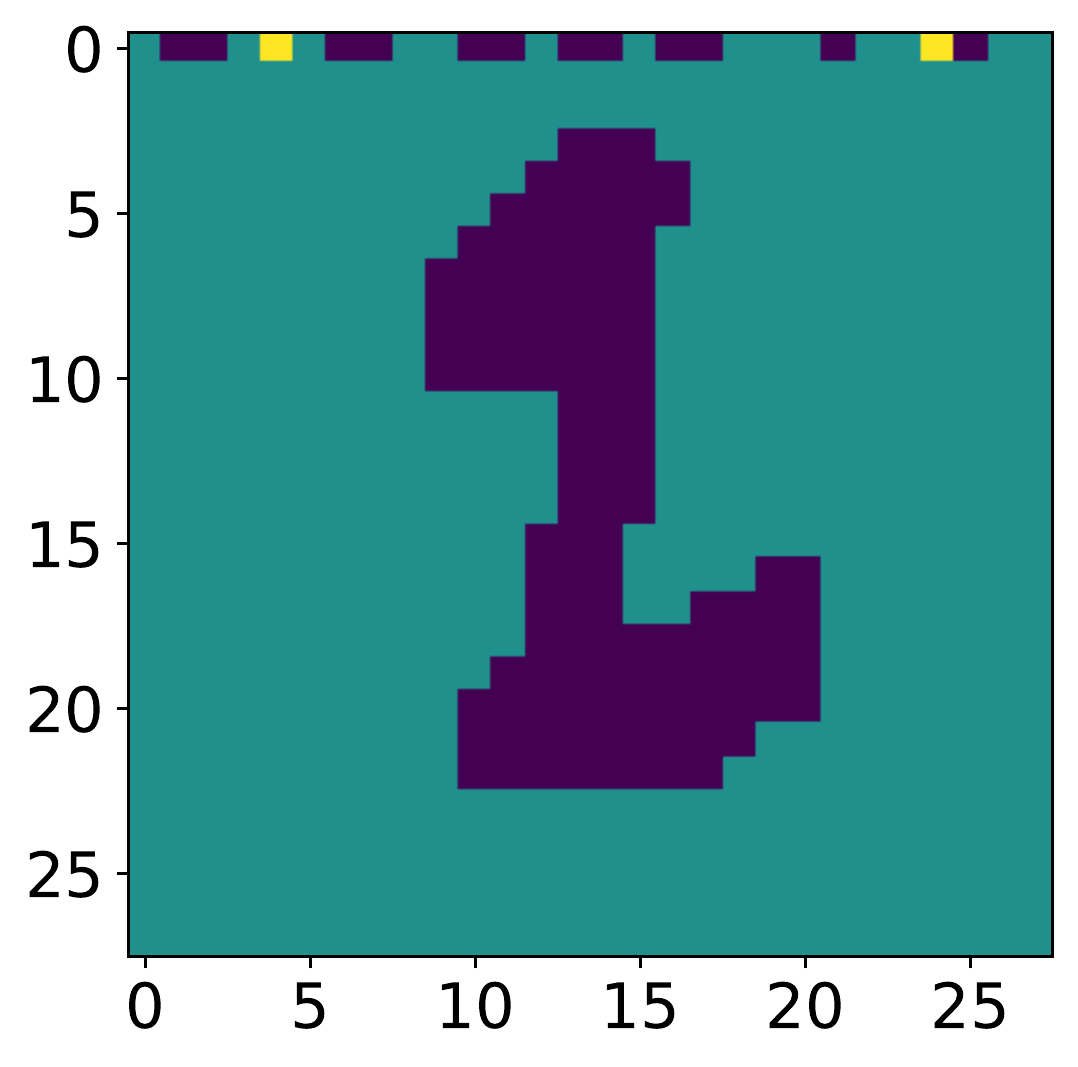}
		\end{minipage}	
	}
\vspace*{-3mm}
	\caption{Graphic representation of essential features and PI-explanations}\label{fig:interpret}
\vspace*{-4mm}
\end{figure}

\smallskip\noindent\textbf{Essential features.}
For the input region given by the Hamming distance $r=4$, we compute two sets of essential features
for the inputs $\ml(G_2^{out})$ and $\ml(G_5^{out})$, i.e., the adversarial examples in the region $R(\bs{u},4)$ that are misclassified into the classes
$2$ and $5$ respectively. The essential features are depicted in Figures~\ref{fig:Ess1to2} and~\ref{fig:Ess1to5},
where black (resp. blue) color means that the value of the corresponding pixel is 1 (resp. 0),
and yellow color means that the value of the corresponding pixel can take arbitrary values.
Figure~\ref{fig:Ess1to2} (resp. Figure~\ref{fig:Ess1to5}) indicates that the inputs $\ml(G_2^{out})$  (resp. $\ml(G_5^{out})$) must
agree on these black- and blue-colored pixels.

\smallskip
\noindent\textbf{PI-explanations.} For demonstration, we assume that the input region is given by the fixed set of indices  $I=\{1,2,\cdots,28\}$
which denotes the first row of pixels of $28\times 28$ images.
We compute two PI-explanations of the inputs $\ml(G_2^{out})$ and $\ml(G_5^{out})$.
The PI-explanations are depicted in Figures~\ref{fig:PI1to2} and~\ref{fig:PI1to5}.
Figure~\ref{fig:PI1to2} (resp. Figure~\ref{fig:PI1to5}) suggests that, by the definition of the PI-explanation, all the images in the region $R(\bs{u},I)$ obtained by assigning
arbitrary values to the yellow-colored pixels are always misclassified into the class
2 (resp. class 5), while changing one black-colored or blue-colored pixel
would change the predication result since a PI-explanation is a minimal set of literals.

\begin{table}[t]
	\centering\setlength{\tabcolsep}{2pt}
	\caption{Maximal safe Hamming distance}\label{tab:SD}
	\begin{adjustbox}{width=1\textwidth,center}
		\begin{tabular}{  c | c r|c  r | c r |c  r  | c r |c  r |c r |c  r }
			\toprule
			\multirow{3}*{\textbf{Image}} & \multicolumn{4}{c|}{P7} & \multicolumn{4}{c|}{P8} & \multicolumn{4}{c|}{ P9}& \multicolumn{4}{c}{ P11} \\
			\cline{2-17}
			~ & \multicolumn{2}{c|}{$\epsilon=0$} & \multicolumn{2}{c|}{$\epsilon=0.03$}  & \multicolumn{2}{c|}{$\epsilon=0$} & \multicolumn{2}{c|}{$\epsilon=0.03$}  & \multicolumn{2}{c|}{$\epsilon=0$} & \multicolumn{2}{c|}{$\epsilon=0.03$}  & \multicolumn{2}{c|}{$\epsilon=0$} & \multicolumn{2}{c}{$\epsilon=0.03$}  \\
			\cline{2-17}
			& SD & Time(s)  & SD &Time(s)  & SD &Time(s)  & SD &Time(s)& SD & Time(s)  & SD &Time(s)  & SD &Time(s)  & SD &Time(s) \\ \midrule	
			\rowcolor{gray!20}\cellcolor{white}0 & 1 &15.09&\bf{4}&10,845&2&0.51&\bf{6}&Timeout&3&746.15&3&737.96&6&29.69&6&29.28  \\	
			1 & -1& 19.96&-1&19.13&-1&2.84&-1&2.97&0&155.50 &0&155.09&0&6.49&0&6.11   \\
			\rowcolor{gray!20}\cellcolor{white}2 &2&13.25&\bf{3}&422.04&0&0.46&0&0.50&1&37.50 &\bf{4}&14,127&6&11,334&6&11,437   \\
			3 & 0&21.39&0&20.94&-1&1.92&-1&2.08&0&41.04&0&40.49&6&8,323.1&6&8,088.3   \\
			\rowcolor{gray!20}\cellcolor{white}4 &3&426.81&\bf{5}&OOM&-1&2.41&-1&2.61&2&8.08&\bf{5}&OOM&6&30.85&6&30.74   \\
			5 &-1&15.60&-1&15.92&-1&0.68&-1&0.74&-1&22.54&-1&21.54&-1&7.03&-1&6.72  \\
			\rowcolor{gray!20}\cellcolor{white}6 &4&7,990.6&\bf{5}&OOM&3&5.69&\bf{4}&198.26&1&57.37&\bf{4}&Timeout&6&44.57&6&45.12   \\
			7&-1&16.08&-1&15.90&-1&2.49&-1&2.52&1&89.49&\bf{4}&Timeout&6&89.38&6&88.39   \\
			\rowcolor{gray!20}\cellcolor{white}8 &-1&19.02&-1&19.28&-1&1.71&-1&1.80&-1&80.16&-1&79.91&6&43.95&6&43.30 \\
			9 &0&26.82&0&27.69&0&5.09&\bf{1}&5.39&-1&109.04&-1&107.24&6&338.73&6&327.48   \\
			\bottomrule
		\end{tabular}
	\end{adjustbox}
	\vspace*{-4mm}
\end{table}


%
%

\section{Related Work}\label{sec:relatedwork}
In this section, we discuss the related work to \tool on qualitative/quantitative analysis and interpretability of DNNs. As there is a vast amount of literature regarding these topics, we will only discuss the most related ones.

\smallskip
\noindent
\textbf{Qualitative analysis of DNNs.} For the verification of real-numbered DNNs, we broadly classify the existing approaches into three categories: (1) constraint solving based, (2) optimization-based, and (3) program analysis based.

The first class of approaches represents the early efforts which
reduce to constraint solving. 
Pulina and Tacchella~\cite{PT10} verified whether the output of the DNN is within an interval
by reducing to the satisfiability checking of a Boolean combination of linear arithmetic constraints
via SMT solvers. Spurious adversarial examples can trigger refinements and retraining of DNNs.
Katz et al.~\cite{KBDJK17} and Ehlers~\cite{Ehl17} independently implemented two SMT solvers, Reluplex and Planet,
for verifying properties of DNNs that are expressible with respective constraints.
Recently, Reluplex was re-implemented in a new framework Marabou~\cite{KatzHIJLLSTWZDK19} with significant improvements.

%

For the second class of approaches which reduce to an optimization
problem, Lomuscio and Maganti~\cite{LomuscioM17}
verified whether some output is reachable from a given input region
by reducing to mixed-integer linear programming (MILP)
via optimization solvers.
To speed up DNN verification via MILP solving,
Cheng et al.~\cite{ChengNR17} proposed heuristics
for MILP encoding and parallelization of MILP solvers.
Dutta et al.~\cite{DuttaJST18} proposed an algorithm to estimate
the output region for a given input region. The algorithm iterates between a global search with MILP solving
and a local search with gradient descent.
Tjeng et al.~\cite{TXT18} proposed a tighter formulation for non-linearities in MILP
and 
methods to improve performance. Recently,
Bunel et al.~\cite{BunelLTTKK20} presented a branch and bound
algorithm (BaB) to verify DNNs on properties expressible in Boolean formulas over linear inequalities.
They claimed that both previous SAT/SMT and MILP-based approaches
are its special cases.
Convex optimization  has also been used to verify DNNs
with over-approximations~\cite{WongK18,DvijothamSGMK18,XiangTJ18}.

For the third class,  researchers have adapted various methods from  traditional static analysis to DNNs. 
A typical example is to use abstract interpretation,
possibly aided with a refinement procedure
to tighten approximations~\cite{GMDTCV18,SGMPV18,SGPV19,LiLYCHZ19,AndersonPDC19,SinghGPV19,LiLHY0ZXH20,YLLHWSXZ20,TranBXJ20,TranLMYNXJ19}.  
These methods vary in the abstract domain (e.g., box, zonotope, polytope, and star-set), efficiency, precision, and activation functions.
(Remark that \cite{SGMPV18,SGPV19} considered floating-points instead of real numbers.)
Another type is to compute convergent output bounds by exploring
neural networks layer-by-layer.
Huang et al.~\cite{HKWW17} proposed an exhaustive search algorithm with an SMT-based refinement.
Later, the search problem was solved via Monte-Carlo tree search~\cite{WickerHK18,WuWRHK20}.
Xiang et al.~\cite{WengZCSHDBD18} proposed to approximate the bounds based on the linear
approximations for the neurons and  Lipschitz constants~\cite{HeinA17}.
Wang et al.~\cite{WPWYJ18} presented symbolic interval analysis
to tighten approximations.
Recently, two abstraction-based frameworks have been proposed~\cite{AshokHKM20,ElboherGK20} which aim
to reduce the size of DNNs, making them more amenable to verification.




Existing techniques for 
quantized DNNs are mostly based on constraint solving, in particular, SAT/SMT solving.
SAT-based approaches transform BNNs into Boolean formulas, where
SAT solving is harnessed~\cite{narodytska2018verifying,neuronFactor18,KorneevNPTBS18,Narodytska18}.
Following this line, verification of three-valued BNNs~\cite{NarodytskaZGW20,JiaR20} and quantized DNNs with multiple bits~\cite{BaranowskiHLNR20,GiacobbeHL20,henzinger2020scalable} were also studied.
Very recently, the SMT-based framework Marabou for real-numbered DNNs~\cite{KatzHIJLLSTWZDK19}
has been extended to support 
BNNs~\cite{AWBK20}.


\smallskip
\noindent
\textbf{Quantitative analysis of DNNs.} Comparing to the qualitative analysis, the quantitative analysis of neural networks is currently very limited.
Two sampling-based approaches were proposed to certify the robustness of adversarial examples~\cite{WebbRTK19,BCMS20},
which require only blackbox access to the models, hence can be applied
on both DNNs and BNNs.
Yang et al.~\cite{YLLHWSXZ20} proposed a spurious region-guided refinement approach for real-numbered DNN verification.
The quantitative robustness verification is achieved by over-approximating
the Lebesgue measure of the spurious regions. The authors claimed that it is the first work to quantitative robustness verification of DNNs
with  soundness guarantee.

Following the SAT-based qualitative analysis of BNNs~\cite{narodytska2018verifying,Narodytska18},
SAT-based quantitative analysis approaches were proposed~\cite{baluta2019quantitative,NarodytskaSMIM19,GBM20}
for verifying robustness and fairness, and assessing heuristic-based explanations of BNNs.
In particular, approximate SAT model-counting solvers are utilized.
As demonstrated in Section~\ref{sec:evaluation}, our BDD-based approach is considerably more accurate and efficient
than the SAT-based one~\cite{baluta2019quantitative}.
In general, we remark that
the BDD construction is computationally expensive but the follow-up analysis 
is often much more efficient,
while the SAT encoding is efficient (polynomial-time) but $\sharp$SAT queries are often computationally expensive
($\sharp$P-hard). The computational cost of our approach is
more dependent on the number of neurons per linear layer but less 
on the number of layers,
while the computational cost of the SAT-based approach~\cite{baluta2019quantitative} 
is 
dependent on both of them.

Shih et al.~\cite{ddlearning19B} proposed a BDD-based
approach to tackle BNNs, similar to our work, in spirit.
In this BDD learning-based approach,
membership queries are implemented by querying the BDD for each input, equivalence queries are implemented by transforming the BDD and BNN to two Boolean formulas,
and checking the equivalence of two Boolean formulas under the input region (in a Boolean formula) via SAT solving.
This construction requires $n$ equivalence queries
and $6n^2+n\cdot \log (m)$ membership queries,
where $n$ (resp. $m$) is the number of nodes (resp. variables) in the final
BDD.
Compared with this approach, our approach is able to handle much larger BNNs than theirs.

\smallskip
\noindent
\textbf{Interpretability of DNNs.}
Though interpretability of DNNs is crucial for
explaining predictions, it is very challenging
to tackle due to the blackbox nature of DNNs.
There is a large body of work on the interpretability of DNNs (cf. \cite{HuangKRSSTWY20,MCB20} for a survey).
Almost all the existing approaches are heuristic-based
and restricted to finding explanations that are local in an input region.
Some of them tackle the interpretability of DNNs by
learning an interpretable model, such as
binary decision trees~\cite{FrosstH17,ZhangYMW19} and finite-state automata~\cite{WeissGY18}.
In contrast to ours, they target at DNNs and only approximate the original model in the input region.
The BDD-based approach~\cite{ddlearning19B} mentioned above has been used to
compute PI-explanation while essential features were not considered therein.

\section{Conclusion}\label{sec:conc}
In this paper, we have proposed a novel BDD-based framework for the quantitative verification 
of BNNs.
We implemented our framework as a prototype tool \tool and conducted extensive experiments on 12 BNN models with varying
sizes and input regions. Experimental results demonstrated that \tool is more scalable than the existing BDD-learning based approach,
and significantly efficient and accurate than the existing SAT-based approach \npaq.
This work represents the first, but a key step of the long-term program to develop an efficient and
scalable BDD-based quantitative analysis framework for BNNs.  



%
%

\appendix
\section{Appendix}

%
%

\begin{table}[t]\setlength{\tabcolsep}{6pt}\renewcommand\arraystretch{1.1}
	\caption{Some basic BDD operations, where $op\in \{\AND,\OR,\XOR,\XNOR\}$} \label{tab:operationBDD}
	\centering
 \scalebox{0.9}{  \begin{tabular}{cc|cc|cc}\toprule
      Operation  & Description &Operation  & Description &Operation  & Description \\ \rowcolor{gray!20}
      $v=\VAR(x)$ & $f_v(x)=x$ &$\EXISTS(v,X)$   & $\exists X.f_v$ & $\SATALL(v)$ & $\SATALL(f_v)$ \\
      $v=\CONST(1)$ & $f_v=1$& $\NOT(v)$  & $\neg f_v$  &$\COMPOSE(v,v')$  & $f_{v}~\circ~ f_{v'}$  \\ \rowcolor{gray!20}    
      $v=\CONST(0)$ & $f_v=0$ & $\APPLY(v,v',op)$ & $f_{v}~op~ f_{v'}$  & $\ITE(x,v,v')$ &$(x\wedge v)\vee (\neg x\wedge v')$ \\
\bottomrule
   \end{tabular}}
\end{table}

\subsection{The Basic BDD Operations}\label{sec:bddoper}
Table~\ref{tab:operationBDD} provides the BDD operations used in this work.
We denote by $op(v,v')$ the operation $\APPLY(v,v',op)$.

\subsection{Proof of Proposition~\ref{prop:inblk2cc}}

Consider the internal block $t_i:\bool^{n_i}\rightarrow \bool^{n_{i+1}}$ for $i\in[d]$.
Since $t_i=(t_i^{bin} \circ t_i^{bn} \circ t_i^{lin})$,
 for every $j\in[n_{i+1}]$ and $\bs{x} \in \bool^{n_i}$,
\[\begin{array}{rl}
 t_{i\downarrow j}(\bs{x}) & = t_i^{bin}( t_i^{bn}(\langle \bs{x}, \bs{W}_{:,j}\rangle+\bs{b}_j)) \\
                           & =t_i^{bin}(t_i^{bn}(\sum_{k=1}^{n_i} \bs{x}_k\cdot\bs{W}_{k,j}+\bs{b}_j))\\
                           & = t_i^{bin}(\alpha_j\cdot(\frac{\sum_{k=1}^{n_i} \bs{x}_k\cdot\bs{W}_{k,j}+\bs{b}_j-\mu_j}{\sigma_j})+\gamma_j) \\
                           & = \left\{\begin{array}{lr}
	+1, \qquad  \text{if}~~ \alpha_j\cdot(\frac{\sum_{k=1}^{n_i} \bs{x}_k\cdot\bs{W}_{k,j}+\bs{b}_j-\mu_j}{\sigma_j})+\gamma_j\geq 0; \\
	-1, \qquad  \text{otherwise}.\\
	\end{array}\right.
\end{array}\]
where $\bs{W}_{k,j}$, $\bs{b}_j$, $\mu_j$, $\alpha_j$, $\sigma_j$, $\gamma_j$ for $k\in [n_i]$ are constants.

%
Therefore, for every $\bs{x} \in \stdbool^{n_i}$, we have:
\[\begin{array}{rl}
t_{i\downarrow j}^{(b)}(\bs{x}) & = t_i^{bin}( t_i^{bn}(\langle 2\bs{x}-\bs{1}, \bs{W}_{:,j}\rangle+\bs{b}_j)),\\
 & = \left\{\begin{array}{lr}
	+1, \qquad  \text{if}~~ \alpha_j\cdot(\frac{\sum_{k=1}^{n_i} (2\bs{x}_k-1)\cdot\bs{W}_{k,j}+\bs{b}_j-\mu_j}{\sigma_j})+\gamma_j\geq 0; \\
	-1, \qquad  \text{otherwise}.\\
	\end{array}\right.
\end{array}\]

Moreover, the constraint $\alpha_j\cdot(\frac{\sum_{k=1}^{n_i} (2\bs{x}_k-1)\cdot\bs{W}_{k,j}+\bs{b}_j-\mu_j}{\sigma_j})+\gamma_j\geq 0$ can be rewritten as  the
cardinality constraint (note: $-\bs{x}_k$ is replaced by $\neg \bs{x}_k-1$)
\[C_{i,j}\triangleq \left\{
\begin{array}{lr}
  \sum_{k=1}^{n_i}\ell_k \geq \lceil \frac{1}{2}\cdot (n_i+\mu_j-\bs{b}_j-\frac{\gamma_j\cdot \sigma_j}{\alpha_j}) \rceil, & \text{if} \ \alpha_j>0;\\
  1, & \text{if} \ \alpha_j=0\wedge \gamma_j\geq 0;\\
  0, & \text{if} \ \alpha_j=0\wedge \gamma_j< 0;\\
  \sum_{k=1}^{n_i}\neg\ell_k \geq \lceil \frac{1}{2}\cdot (n_i-\mu_j+\bs{b}_j+\frac{\gamma_j\cdot \sigma_j}{\alpha_j}) \rceil,   & \text{if} \ \alpha_j<0;
\end{array}\right.\]
where for every $k\in [n_i]$, $\ell_k$ is $\bs{x}_k$ if $\bs{W}_{k,j}=+1$, and $\ell_k$ is $\neg\bs{x}_k$ if $\bs{W}_{k,j}=-1$.

Thus, we get that $t_{i\downarrow j}^{(b)}\Leftrightarrow C_{i,j}$.

\subsection{Proof of Proposition~\ref{prop:outblk2cc}}

For the output block $t_{d+1}:\bool^{n_{d+1}}\rightarrow \stdbool^{s}$,
since $t_{d+1}=t_{d+1}^{am} \circ t_{d+1}^{lin}$, then for every $j\in[s]$ and $\bs{x}\in \bool^{n_{d+1}}$,
 \[ \begin{array}{rl} t_{d+1\downarrow j}(\bs{x})=1 & \text{iff} \
 \left( \begin{array}{c}
 \forall j'\in [j-1].\sum_{k=1}^{n_{d+1}} \bs{x}_k\cdot\bs{W}_{k,j}+\bs{b}_j>\sum_{k=1}^{n_{d+1}} \bs{x}_k\cdot\bs{W}_{k,j'}+\bs{b}_j'   \\
 \text{and}\\
 \forall j'\in \{j+1,\cdots,s\}.\sum_{k=1}^{n_{d+1}} \bs{x}_k\cdot\bs{W}_{k,j}+\bs{b}_j\geq \sum_{k=1}^{n_{d+1}} \bs{x}_k\cdot\bs{W}_{k,j'}+\bs{b}_j'
\end{array}\right)
\end{array}.\]

Therefore, for every $\bs{x}\in \stdbool^{n_{d+1}}$, we have:
\[ t_{d+1\downarrow j}^{(b)}(\bs{x})=1 \ \text{iff} \
 \left( \begin{array}{c}
 \forall j'\in [j-1].\sum_{k=1}^{n_{d+1}} (2\bs{x}_k-1)\cdot (\bs{W}_{k,j}-\bs{W}_{k,j'})>\bs{b}_j'-\bs{b}_j  \\
 \text{and}\\
 \forall j'\in \{j+1,\cdots,s\}.\sum_{k=1}^{n_{d+1}} (2\bs{x}_k-1)\cdot (\bs{W}_{k,j}-\bs{W}_{k,j'})\geq \bs{b}_j'-\bs{b}_j
\end{array}\right),\]
where the latter holds iff
\[ \left( \begin{array}{c}
 \forall j'\in [j-1].\sum_{k=1}^{n_{d+1}} \bs{x}_k\cdot \frac{\bs{W}_{k,j}-\bs{W}_{k,j'}}{2}>\frac{1}{4}(\bs{b}_j'-\bs{b}_j+ \sum_{k=1}^{n_{d+1}} (\bs{W}_{k,j}-\bs{W}_{k,j'}))  \\
 \text{and}\\
 \forall j'\in \{j+1,\cdots,s\}.\sum_{k=1}^{n_{d+1}} \bs{x}_k\cdot \frac{\bs{W}_{k,j}-\bs{W}_{k,j'}}{2}\geq \frac{1}{4}(\bs{b}_j'-\bs{b}_j + \sum_{k=1}^{n_{d+1}} (\bs{W}_{k,j}-\bs{W}_{k,j'}))
\end{array}\right).
\]

For every $j'\in [j-1]$, $\sum_{k=1}^{n_{d+1}} \bs{x}_k\cdot \frac{\bs{W}_{k,j}-\bs{W}_{k,j'}}{2}>\frac{1}{4}(\bs{b}_j'-\bs{b}_j+ \sum_{k=1}^{n_{d+1}} (\bs{W}_{k,j}-\bs{W}_{k,j'})) $
can be rewritten as the cardinality constraint
\[C_{d+1,j'}\triangleq \left\{
\begin{array}{l}
\sum_{k=1}^{n_{d+1}} \ell_{d+1,k} \geq \frac{1}{4}(\bs{b}_j'-\bs{b}_j+ \sum_{k=1}^{n_{d+1}} (\bs{W}_{k,j}-\bs{W}_{k,j'}))+1+\sharp{\tt Neg}, ~~~~~~~~~~~~~~~~~~~~~~~   ~~~~~~~~~~~~~~~~~  \\
\hfill  ~~~~~~~~~~~~~~~~~~~~~~~~~~~~~~~~ \text{if }  \frac{1}{4}(\bs{b}_j'-\bs{b}_j+ \sum_{k=1}^{n_{d+1}} (\bs{W}_{k,j}-\bs{W}_{k,j'})) \ \text{is an integer};\\ \\
\sum_{k=1}^{n_{d+1}} \ell_{d+1,k} \geq \lceil\frac{1}{4}(\bs{b}_j'-\bs{b}_j+ \sum_{k=1}^{n_{d+1}} (\bs{W}_{k,j}-\bs{W}_{k,j'}))\rceil+\sharp {\tt Neg},  \hfill \text{otherwise}; \\
\end{array}\right.\]
where $\sharp {\tt Neg}=|\{k\in [n_{d+1}] \mid \bs{W}_{k,j}-\bs{W}_{k,j'}=-2\}|$, $\ell_{d+1,k}$ is $\bs{x}_{d+1,k}$ if $\bs{W}_{k,j}-\bs{W}_{k,j'}=+2$,
$\ell_{d+1,k}$ is  $\neg \bs{x}_{d+1,k}$ if $\bs{W}_{k,j}-\bs{W}_{k,j'}=-2$,
and $\ell_{d+1,k}$ is $0$ if $\bs{W}_{k,j}-\bs{W}_{k,j'}=0$.

Similarly, $\sum_{k=1}^{n_{d+1}} \bs{x}_k\cdot \frac{\bs{W}_{k,j}-\bs{W}_{k,j'}}{2}\geq \frac{1}{4}(\bs{b}_j'-\bs{b}_j + \sum_{k=1}^{n_{d+1}} (\bs{W}_{k,j}-\bs{W}_{k,j'}))$ for each $j'\in \{j+1,\cdots,s\}$
can be rewritten as the cardinality constraint
\[C_{d+1,j'}\triangleq \sum_{k=1}^{n_{d+1}} \ell_{d+1,k} \geq \lceil\frac{1}{4}(\bs{b}_j'-\bs{b}_j+ \sum_{k=1}^{n_{d+1}} (\bs{W}_{k,j}-\bs{W}_{k,j'}))\rceil+\sharp {\tt Neg},\]
where $\sharp {\tt Neg}=|\{k\in [n_{d+1}] \mid \bs{W}_{k,j}-\bs{W}_{k,j'}=-2\}|$, $\ell_{d+1,k}$ is $\bs{x}_{d+1,k}$ if $\bs{W}_{k,j}-\bs{W}_{k,j'}=+2$,
$\ell_{d+1,k}$ is  $\neg \bs{x}_{d+1,k}$ if $\bs{W}_{k,j}-\bs{W}_{k,j'}=-2$,
and $\ell_{d+1,k}$ is $0$ if $\bs{W}_{k,j}-\bs{W}_{k,j'}=0$.

Thus, we get that $t_{d+1\downarrow j}^{(b)}\Leftrightarrow  C_{d+1,1}\wedge\cdots \wedge C_{d+1,j-1}\wedge C_{d+1,j+1}\wedge\cdots\wedge C_{d+1,s}$.

\subsection{Explanation of Algorithm~\ref{alg:dnn2bdd}}\label{sec:explanation}
Given a BNN $\mb=(t_1,\cdots,t_d,t_{d+1})$ with $s$ output classes and an input region $R(\bs{u},\tau)$,
Algorithm~\ref{alg:dnn2bdd} outputs the BDDs $(\mg_i^{out})_{i\in[s]}$,
encoding the input-output relation of the BNN $\mb$
w.r.t. the input region $R(\bs{u},\tau)$.

In detail, it first builds the BDD representation $\mg^{in}_{\bs{u},\tau}$ of the input region $R(\bs{u},\tau)$
and the cardinality constraints from BNN $\mb^{(b)}$ (Line 1).

\smallskip
\noindent
{\bf The first for-loop}. It builds a BDD encoding the input-output relation of the entire internal blocks w.r.t. $\mg^{in}_{\bs{u},\tau}$.
It first invokes the procedure {\sc Block2BDD}$(t_i^{b},\mg^{in},i)$
to build a BDD $\mg'$ encoding the input-output relation of the $i$-{th} block $t_{i}^{b}$ w.r.t.
$\ml(\mg^{in})$ (Line 4). $\mg^{in}$ is the set of feasible inputs of the block
$t_{i}^{b}$, which is also the set of feasible outputs of the $(i-1)$-{th} block
$t_{i-1}^{b}$ (the input region $\mg^{in}_{\bs{u},\tau}$ when $i=1$).
By doing so, we have: \[\ml(\mg')=\{(\bs{x}^i,\bs{x}^{i+1})\in \ml(\mg^{in})\times \stdbool^{n_i+1} \mid t_i^{(b)}(\bs{x}^i)=\bs{x}^{i+1}\}.\]
From the BDD $\mg'$, we compute the feasible outputs $\mg^{in}$ of the block $t_i^{(b)}$ by existentially quantifying
all the input variables $\bs{x}^i$ of the block $t_i^{(b)}$ (Line 5). The BDD $\mg^{in}$  serves as the set of feasible inputs of the block $t_{i+1}^{(b)}$ at the next iteration.

We next assign $\mg'$ to $\mg$ if the current block is the first internal block (i.e., $i=1$), otherwise we compute the relational product 
of $\mg'$ and $\mg$, the resulting BDD $\mg$ encodes the input-output relation of the first $i$ internal blocks w.r.t. $\mg^{in}_{\bs{u},\tau}$ (Line 6),
namely,
\begin{center}
$\ml(\mg)=\{(\bs{x}^1,\bs{x}^{i+1})\in \ml(\mg^{in}_{\bs{u},\tau})\times \stdbool^{n_i+1} \mid (t_i^{(b)}\circ\cdots \circ t_1^{(b)})(\bs{x}^1)=\bs{x}^{i+1}\}$.
\end{center}

Furthermore,
\begin{center}
$\ml(\mg^{in})=\{\bs{x}^{i+1}\in \stdbool^{n_i+1} \mid \exists \bs{x}^1\in \ml(\mg^{in}_{\bs{u},\tau}). (\bs{x}^1,\bs{x}^{i+1})\in \ml(\mg)\}$.
\end{center}
At the end of the first for-loop, we obtain the BDD $\mg$ encoding
the input-output relation of the entire internal blocks and its feasible outputs $\mg^{in}$ w.r.t. $\mg^{in}_{\bs{u},\tau}$,
namely,
\[\begin{array}{l}
  \ml(\mg)=\{(\bs{x}^1,\bs{x}^{d+1})\in \ml(\mg^{in}_{\bs{u},\tau})\times \stdbool^{n_{d+1}} \mid (t_d^{(b)}\circ\cdots \circ t_1^{(b)})(\bs{x}^1)=\bs{x}^{d+1}\},\\
 \ml(\mg^{in})=\{\bs{x}^{d+1}\in \stdbool^{n_{d+1}} \mid \exists \bs{x}^1\in \ml(\mg^{in}_{\bs{u},\tau}).
(\bs{x}^1,\bs{x}^{d+1})\in\ml(\mg)\}. \\
\end{array}\]

%


\smallskip
\noindent
{\bf The second for-loop}. It builds the BDDs $(\mg_i^{out})_{i\in[s]}$, each of which
encodes the input-output relation of the entire BNN
and a class $i\in[s]$ w.r.t. $\mg^{in}_{\bs{u},\tau}$. For each  $i\in[s]$,
it first builds a BDD $\mg_i$ encoding the input-output relation
of the output block for the class $i$ by invoking {\sc Block2BDD}$(t_{d+1\downarrow i}^{(b)},\mg^{in},d+1)$ (Line 8).
By computing the relational product of the BDDs $\mg_i$ and $G$, we obtain
the BDD $\mg_i^{out}$. 
Recall that the BDD $G$
encodes the input-output relation of the entire internal blocks  w.r.t. the input region $\mg^{in}_{\bs{u},\tau}$.
Thus, an input $\bs{x}\in R(\bs{u},\tau)$ is classified into the class $i$ by the BNN $\mb$ iff
$\bs{x}^{(b)}\in\ml(G_i^{out})$.

\smallskip
\noindent
{\bf The procedure {\sc Block2BDD}}. It receives the cardinality
constraints $\{C_m,\cdots,C_n\}$ (note that indices matter), a BDD $\mg^{in}$ encoding feasible inputs
and the block index $i$ as inputs, and returns a BDD $\mg$.
\begin{itemize}
  \item If $i=d+1$, namely, the cardinality
constraints $\{C_m,\cdots,C_n\}$ are from the output block,
the resulting BDD $\mg$ encodes the subset of $\mg^{in}_{\bs{u},\tau}$ that satisfy all the cardinality
constraints $\{C_m,\cdots,C_n\}$.
\item If $i\neq d+1$, then
the BDD $\mg$ encodes the input-output relation of the Boolean function $f_{m,n}$ such that
for every $\bs{x}^{i}\in \ml(G^{in})$, $f_{m,n}(\bs{x}^{i})$
is the truth vector of the cardinality
constraints $\{C_m,\cdots,C_n\}$ under the valuation $\bs{x}^{i}$.
When $m=1$ and $n=n_{i+1}$, $f_{m,n}$ is the same as $t_i^{(b)}$, hence
\[\ml(\mg)=\{\bs{x}^i\times \bs{x}^{i+1}\in \mg^{in}\times \stdbool^{n_{i+1}}\mid t_i^{(d)}(\bs{x}^i)=\bs{x}^{i+1}\}.\]
\end{itemize}
In detail, the procedure {\sc Block2BDD} computes the desired BDD in a binary search fashion.
\begin{itemize}
  \item If $m=n$, it first builds the BDD $G_1$ for the cardinality
constraint $C_n$ (Line 13) such that $\ml(G_1)$ is the set of solutions
of $C_n$. Then it computes the conjunction $G$ of the BDDs $G_1$
and $G^{in}$. 
Recall that  $G^{in}$ is the set of feasible input of the $i$-th block $t_{i}^{(b)}$.
Thus, $\ml(G)\subseteq\ml(G^{in})$ is the set of feasible inputs that satisfy $C_n$.
If $i\neq d+1$, the BDD $G$ is transformed into the BDD $\XNOR(\bs{x}_m^{i+1},\mg)$.
This step encodes the constraint $\bs{x}_m^{i+1}\Leftrightarrow C_n$
in the BDD $\XNOR(\bs{x}_m^{i+1},\mg)$, namely,
for every $(\bs{u},b)\in\ml(\XNOR(\bs{x}_m^{i+1},\mg))$,
$b$ is the truth of $C_n$ under the valuation $\bs{u}$.
Remark that $\bs{x}_m^{i+1}$ is a new Boolean variable introduced into
the BDD $\XNOR(\bs{x}_m^{i+1},\mg)$.

  \item If $m\neq n$, we recursively build the BDDs $G_1$ and $G_2$ for
$\{C_m,\cdots,C_{\lfloor\frac{n-m}{2}\rfloor+m}\}$ and $\{C_{\lfloor\frac{n-m}{2}\rfloor+m+1},\cdots,C_n\}$
and compute the conjunction of $G_1$ and $G_2$.
Thus, if $i=d+1$, $\ml(G)\subseteq \ml(G^{in})$ is the set of all the feasible
inputs that satisfy the cardinality constraints $\{C_m,\cdots,C_n\}$.
If $i\neq d+1$, for every $(\bs{u},\bs{x})\in \ml(G)$,
$\bs{x}$ is the truth vector of the constraints $\{C_m,\cdots,C_n\}$ under the valuation
$\bs{u}$.
\end{itemize}

\begin{figure}[t]
	\centering
	\subfigure{
		\begin{minipage}[b]{0.08\textwidth}
			\includegraphics[width=1\linewidth]{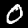}
		\end{minipage}	
	}
	\subfigure{
		\begin{minipage}[b]{0.08\textwidth}
			\includegraphics[width=1\linewidth]{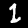}
		\end{minipage}	
	}
	\subfigure{
		\begin{minipage}[b]{0.08\textwidth}
			\includegraphics[width=1\linewidth]{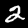}
		\end{minipage}	
	}
	\subfigure{
		\begin{minipage}[b]{0.08\textwidth}
			\includegraphics[width=1\linewidth]{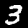}
		\end{minipage}	
	}
	\subfigure{
		\begin{minipage}[b]{0.08\textwidth}
			\includegraphics[width=1\linewidth]{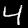}
		\end{minipage}
	}
	\subfigure{
		\begin{minipage}[b]{0.08\textwidth}
			\includegraphics[width=1\linewidth]{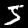}
		\end{minipage}	
	}
	\subfigure{
		\begin{minipage}[b]{0.08\textwidth}
			\includegraphics[width=1\linewidth]{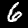}
		\end{minipage}	
	}
	\subfigure{
		\begin{minipage}[b]{0.08\textwidth}
			\includegraphics[width=1\linewidth]{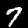}
		\end{minipage}	
	}
	\subfigure{
		\begin{minipage}[b]{0.08\textwidth}
			\includegraphics[width=1\linewidth]{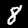}
		\end{minipage}	
	}
	\subfigure{
		\begin{minipage}[b]{0.08\textwidth}
			\includegraphics[width=1\linewidth]{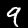}
		\end{minipage}	
	}
	\caption{10 randomly chosen images used to evaluate our approach}\label{fig:784inputs}
\end{figure}

\begin{table}[ht]\setlength{\tabcolsep}{6pt}\renewcommand\arraystretch{1.1}
	\caption{Class distribution of samples on P8 under Hamming distance $r=2,3,4$}\label{tab:P8robustHD}
	\begin{adjustbox}{width=0.95\textwidth,center}
	\begin{tabular}{  c  c | rrrr rrrrrr }
		\toprule
		\multicolumn{2}{c|}{Region} &  0  & 1 & 2 & 3 & 4 & 5 & 6 & 7 & 8 & 9 \\	\midrule
		 & \ 0 \ & 5,051 & 0 & 0 & 0 & 0 & 0 & 0 & 0 &0& 0 \\ \rowcolor{gray!20}
		\cellcolor{white} & \ 1 \ & 0 & 0 & 163 & 4,006 & 0 & 354 & 102 & 0 & 426 & 0 \\
		 & \ 2 \ &0&0&4,193&858&0&0&0&0&0&0\\  \rowcolor{gray!20}
		\cellcolor{white} & \ 3 \  &43&0&3&3,037&0&1,571&0&0&397&0\\
		 & \ 4 \  &0&0&1,969&555&1,286&23&1,140&0&3&75\\  \rowcolor{gray!20}
		\cellcolor{white} & \ 5 \  &0&0&0&4,721&0&293&0&0&37&0\\
		 & \ 6 \ &0&0&0&0&0&0&5,051&0&0&0\\  \rowcolor{gray!20}
		\cellcolor{white} & \ 7 \  &0&0&0&36&0&304&0&2,878&291&1,542\\
		 & \ 8 \  &0&0&0&0&13&3,011&0&0&2,027&0\\  \rowcolor{gray!20}
		\cellcolor{white}\multirow{-10}*{\textbf{r=2}}  & \ 9 \  &4&0&2&0&459&9&51&9&429&4,088\\
		\midrule
		  & \ 0 \ & 166,733 &0&0&0&0&2&16&0&0&0 \\ \rowcolor{gray!20}
		\cellcolor{white}
		 & \ 1 \ &47&25&8,562&119,551&0&7,277&7,074&63&24,152&0 \\
		 & \ 2 \ &635&0&132,938&26,754&0&0&0&0&5,885&539\\\rowcolor{gray!20}
		\cellcolor{white}
		 & \ 3 \  &7,377&0&182&91,683&0&55,560&0&0&11,949&0\\
		 & \ 4 \  &0&1,104&19,304&11,810&87,256&8,446&17,676&3,736&7,798&9,621\\\rowcolor{gray!20}
		\cellcolor{white}
		 & \ 5 \  &0&0&14&142,837&0&16,815&0&10&7,075&0\\
		 & \ 6 \ &0&0&0&0&0&0&166,751&0&0&0\\\rowcolor{gray!20}
		\cellcolor{white}
		 & \ 7 \  &0&0&0&800&11&8,441&0&118,102&17,790&21,607\\
		 & \ 8 \  &0&0&0&0&1,903&55,793&69&257&108,727&2\\\rowcolor{gray!20}
		\cellcolor{white}
		\multirow{-10}*{\textbf{r=3}}  & \ 9 \  &960&0&51&0&10,518&1,636&729&955&17,141&134,761\\
		\midrule
  & \ 0 \ & 4,086,659 & 0 & 0 & 0 & 0 & 548 & 769 & 0 & 0 & 0 \\ \rowcolor{gray!20}
		\cellcolor{white}
		 & \ 1 \ & 18,013 &3,085&239,603&2,925,674&0&361,402&207,126&10,092&322,977&4 \\
		 & \ 2 \ &5,329&0&2,933,180&860,024&0&0&2&0&283,998&5,443\\ \rowcolor{gray!20}
		\cellcolor{white}
		 & \ 3 \  &120,137&0&19,533&2,597,188&0&767,756&63&0&583,299&0\\
		 & \ 4 \  &0&62,337&1,047,340&878,728&949,205&144,075&682,230&35,943&161,122&126,996\\ \rowcolor{gray!20}
		\cellcolor{white}
		 & \ 5 \  &0&0&823&3,471,985&0&490,715&0&360&124,093&0\\
		 & \ 6 \ &68,279&0&1,844&1,248&2&12,892&4,002,988&0&29&694\\ \rowcolor{gray!20}
		\cellcolor{white}
		 & \ 7 \  &6&0&0&62,549&107&218,970&0&2,738,751&282,666&784,927\\
		 & \ 8 \  &1,190&11&9&1&19,604&2,048,007&1,302&8,104&2,008,890&858\\ \rowcolor{gray!20}
		\cellcolor{white}
		\multirow{-10}*{\textbf{r=4}}   & \ 9 \  &31,828&0&3,395&22&411,798&63,901&61,127&75,933&661,904&2,778,068\\
\bottomrule
	\end{tabular}
\end{adjustbox}
\end{table}

\begin{table}[ht]\setlength{\tabcolsep}{0pt}\renewcommand\arraystretch{1.1}
	\caption{Class distribution of samples on P11 under Hamming distance $r=2,3,4$}\label{tab:P11robustHD}
	\begin{adjustbox}{width=1\textwidth,center}
    \begin{tabular}{  c  c | ccccc ccccc }
		\toprule
		\multicolumn{2}{c|}{Region}  &  0  & 1 & 2 & 3 & 4 & 5 & 6 & 7 & 8 & 9 \\
		\midrule
	  & \ 0 \  &307,721&0&0&0&0&0&0&0&0&0\\ \rowcolor{gray!20}
		\cellcolor{white}
		 & \ 1 \  &0&258,705&49,016&0&0&0&0&0&0&0\\
		 & \ 2 \  &0&0&307,721&0&0&0&0&0&0&0\\\rowcolor{gray!20}
		\cellcolor{white}
		 & \ 3 \  &0&0&0&307,721&0&0&0&0&0&0\\
		 & \ 4 \  &0&0&0&0&307,721&0&0&0&0&0\\\rowcolor{gray!20}
		\cellcolor{white}
		 & \ 5 \  &0&0&0&238,346&0&69,375&0&0&0&0\\
		 & \ 6 \  &0&0&0&0&0&0&307,721&0&0&0\\\rowcolor{gray!20}
		\cellcolor{white}
		 & \ 7 \  &0&0&0&0&0&0&0&307,721&0&0\\
		 & \ 8 \  &0&0&0&0&0&0&0&0&307,721&0\\\rowcolor{gray!20}
		\cellcolor{white}
		\multirow{-10}*{\textbf{r=2}}   & \ 9 \  &0&0&0&0&0&0&0&0&0&307,721\\
		\midrule
		  & \ 0 \ &80,315,705&0&0&0&0&0&0&0&0&0\\\rowcolor{gray!20}
		\cellcolor{white}
		 & \ 1 \  &0&63,358,484&16,957,221&0&0&0&0&0&0&0\\
		 & \ 2 \  &0&0&80,315,705&0&0&0&0&0&0&0\\\rowcolor{gray!20}
		\cellcolor{white}
		 & \ 3 \  &0&0&0&80,315,705&0&0&0&0&0&0\\
		 & \ 4 \  &0&0&0&0&80,315,705&0&0&0&0&0\\\rowcolor{gray!20}
		\cellcolor{white}
		 & \ 5 \  &0&79,523&0&54,664,874&0&25,571,308&0&0&0&0\\
		 & \ 6 \  &0&0&0&0&0&0&80,315,705&0&0&0\\\rowcolor{gray!20}
		\cellcolor{white}
		 & \ 7 \  &0&0&0&0&0&0&0&80,315,705&0&0\\
		 & \ 8 \  &0&0&0&0&0&0&0&0&80,315,705&0\\\rowcolor{gray!20}
		\cellcolor{white}
		\multirow{-10}*{\textbf{r=3}}   & \ 9 \  &0&0&0&0&0&0&0&0&0&80,315,705\\
		\midrule
        & \ 0 \  &15,701,874,581&0&0&0&0&0&0&0&0&0\\\rowcolor{gray!20}
		\cellcolor{white}
		 & \ 1 \  &0&12,939,606,785&2,762,267,796&0&0&0&0&0&0&0\\
		 & \ 2 \  &0&0&15,701,874,581&0&0&0&0&0&0&0\\\rowcolor{gray!20}
		\cellcolor{white}
		 & \ 3 \  &0&0&0&15,701,874,581&0&0&0&0&0&0\\
		 & \ 4 \  &0&0&0&0&15,701,874,581&0&0&0&0&0\\\rowcolor{gray!20}
		\cellcolor{white}
		 & \ 5 \  &0&721,872&0&10,938,034,150&0&4,763,118,559&0&0&0&0\\
		 & \ 6 \  &0&0&0&0&0&0&15,701,874,581&0&0&0\\\rowcolor{gray!20}
		\cellcolor{white}
		 & \ 7 \  &0&0&0&0&0&0&0&15,701,874,581&0&0\\
		 & \ 8 \  &0&0&0&0&0&0&0&0&15,701,874,581&0\\\rowcolor{gray!20}
		\cellcolor{white}
		 \multirow{-10}*{\textbf{r=4}}    & \ 9 \  &0&0&0&0&0&0&0&0&0&15,701,874,581\\
\bottomrule
	\end{tabular}
\end{adjustbox}
\end{table}

\subsection{Distributions of Classes on P8 and P11 with Hamming Distance $r=2,3,4$}
Table~\ref{tab:P8robustHD} shows the number of samples for each input region
and  class on P8, while Table~\ref{tab:P11robustHD} shows the number of samples for each input region
and  class on P11, where the input regions are given by Hamming Distance $r=2,3,4$.

\subsection{Results on BNNs under Indices based Input Regions}\label{sec:expFI}
{\bf BDD encoding under fixed indices.}
We evaluate \tool on the BNNs (P5--P12) using the 10 images. In this case, the input regions
are given by the fixed indices $I$ with size ranging from 10 to 30.
The results on average of the images are shown in Table~\ref{tab:locEncFI}.
\tool is able to encode all the BNNs when $|I|\leq 25$.
We can observable similar results as BDD encoding under Hamming distance, namely,
the execution time and number of BDD nodes increase
with the size of $I$.

\begin{table}[t]
	\centering\setlength{\tabcolsep}{5pt}
	\caption{BDD encoding under fixed indices}
	\begin{adjustbox}{width=1\textwidth,center}
		\label{tab:locEncFI}
		\begin{tabular}{c|rr|rr|rr|rr|rr}
			\toprule
			& \multicolumn{2}{c|}{$|I|$=10} &\multicolumn{2}{c|}{$|I|$=15} &\multicolumn{2}{c|}{$|I|$=20} &\multicolumn{2}{c}{$|I|$=25} &\multicolumn{2}{c}{$|I|$=30}\\
			~& Time(s) &$| \mg |$&Time(s)&$| \mg |$&Time(s)&$| \mg |$&Time(s)&$| \mg |$&Time(s)&$| \mg |$\\
			\midrule \rowcolor{gray!20}
			P5 & 0.01 & 1,271 & 0.03 & 10,516 & 0.23 & 69,9901 & 3.87 & 980,733 & 47.80 & 2,852,039  \\ 
			
			P6 & 0.15 & 1,740 & 0.90 & 52,067 & 35.56 & 699,369 & 790.51& 9,890,720 & (3) 21,819&182,893,693 \\ 
			\rowcolor{gray!20}
			P7 & 0.43 & 2,095 & 4.61 & 49,002 & 232.14 & 656,068 & 7,842.9 & 12,661,328 & [10]  & - \\  
			
			P8 & 0.17 & 2,614 & 0.88 & 55,359 & 24.84 & 1,073,596 & 583.95 & 14,629,891 & (2) 16,388& 325,658,126 \\ 
			\rowcolor{gray!20}
			P9 & 1.14 & 2,058 & 17.18 & 59,515 & 665.56 & 875,011 & 15,240 & 15,438,852& [10] &- \\ 
			
			P10 & 0.20 & 2,057 & 1.56 & 52,398 & 47.82 & 895,610 & 1,057.8& 15,075,904 & [3](5) 25,165& 498,593,282 \\ 
			\rowcolor{gray!20}
			P11 & 6.27 & 1,667 & 6.39 & 13,497 & 6.74 & 56,707 & 11.36 & 346,016 & 41.23 & 2,355,023 \\
			
			P12 & 12.55 & 2,674 & 13.16 & 23,139 & 15.19 & 172,897 & 64.66 & 3,692,808 & 847.47 & 30,955,447 \\
			\bottomrule
		\end{tabular}}
	\end{adjustbox}
\end{table}

\smallskip
\noindent
{\bf Robustness verification with fixed indices.}
We illustrate \tool on 2 representative BNNs (P8 and P11), where the input regions
are given by the sets of fixed indices:
$I_8=\{1,2,\cdots,10\}$ for P8
and $I_{11}=\{1,2,\cdots,28\}$ for P11,
corresponding to the first row of pixels of $10\times 10$ and $28\times 28$ images. Each region was successfully verified in less than
1 minute.

\begin{figure}[t]
	\centering
	\subfigure[Results of P8]{\label{fig:p8FI}
	\begin{minipage}[c]{.45\textwidth}
		\centering
		\includegraphics[width=1\textwidth]{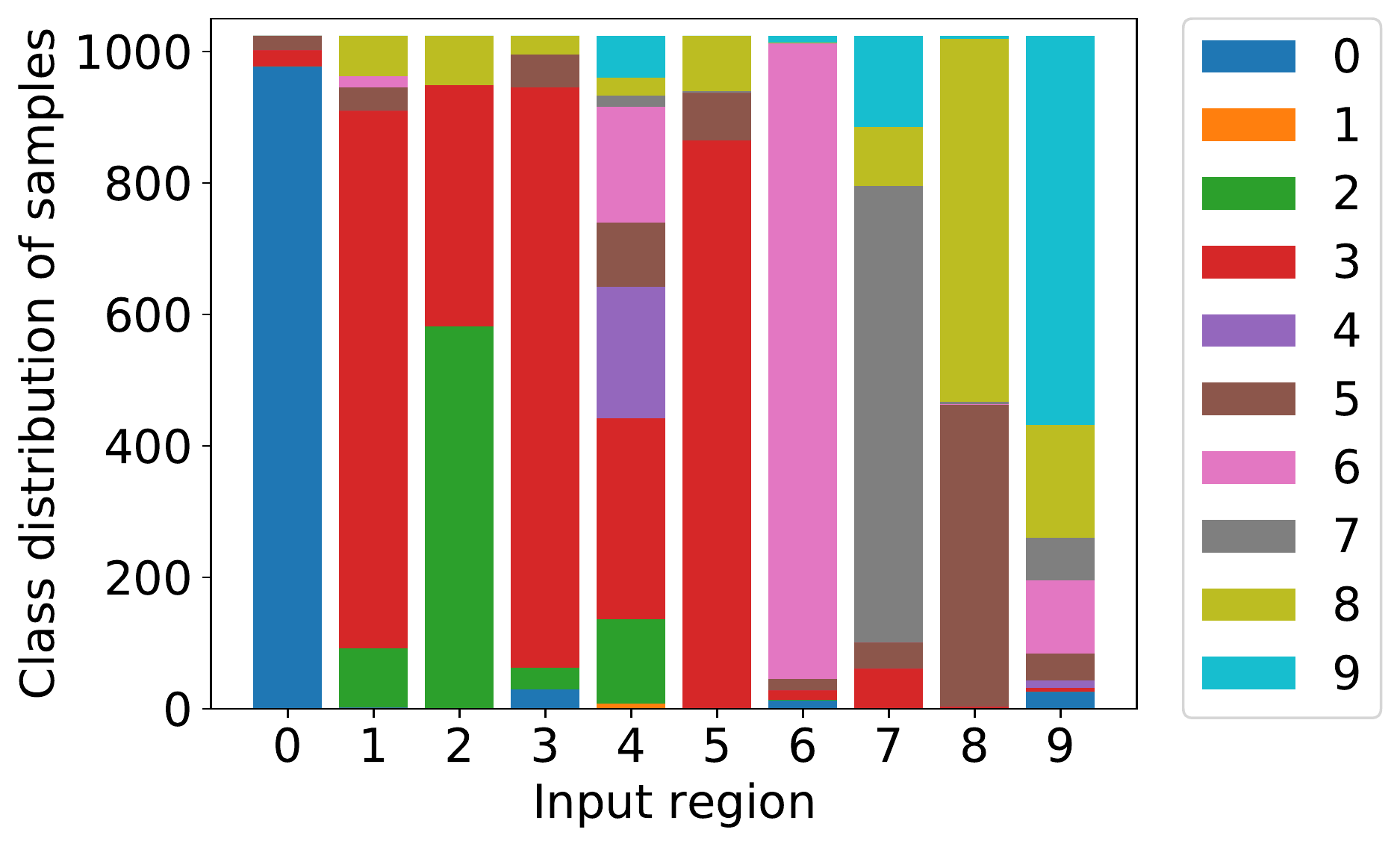}
	\end{minipage}
	}
	\subfigure[Results of P11]{\label{fig:P11FI}
	\begin{minipage}[c]{.5\textwidth}
		\centering
		\includegraphics[width=1\textwidth]{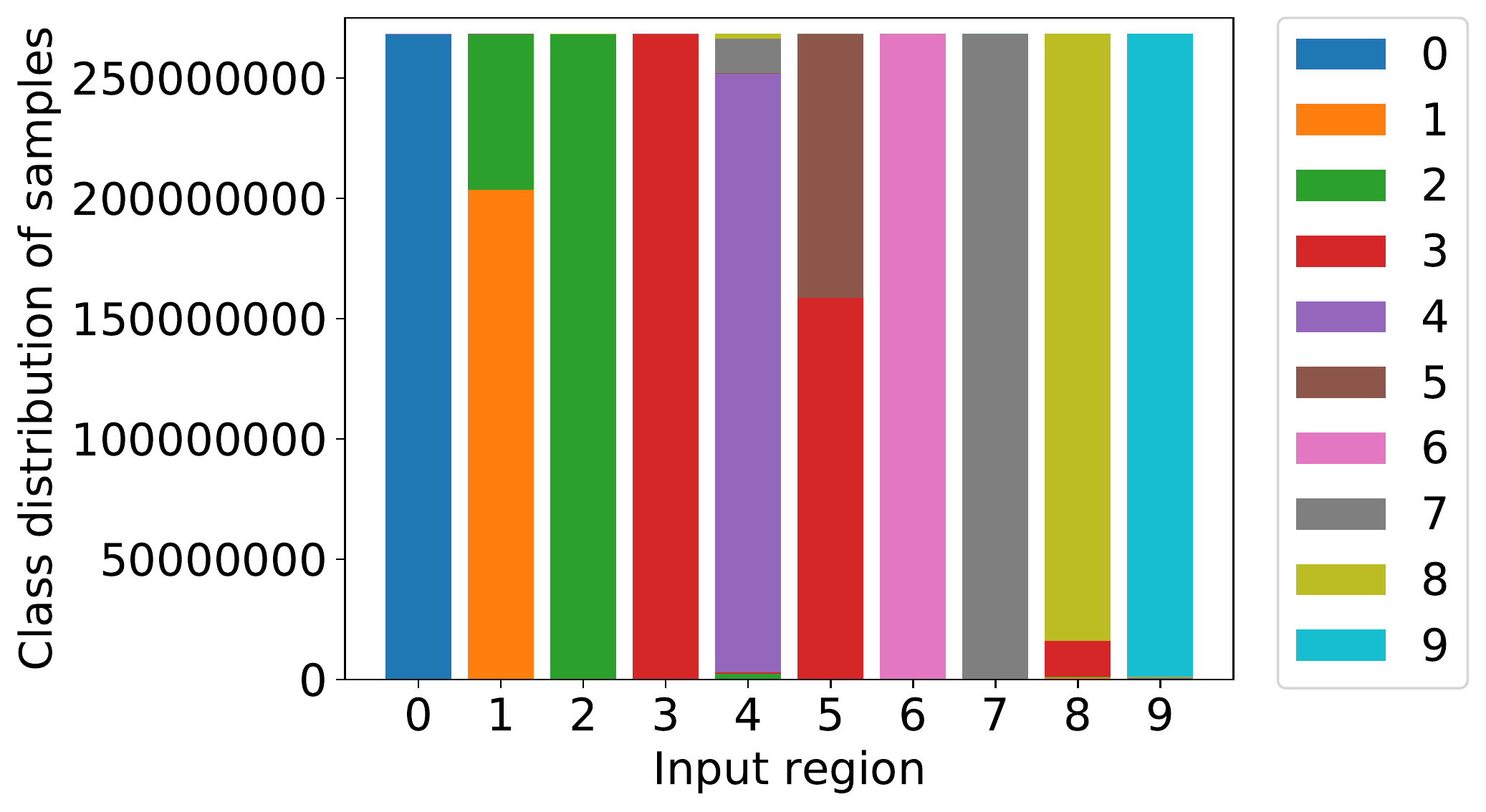}
	\end{minipage}
	}
	\caption{Distribution of classes of robustness verification with fixed indices}
\label{fig:P8P11FI}
\end{figure}

Figure~\ref{fig:P8P11FI} shows the distribution of classes of robustness verification with fixed indices.
Raw data are shown in Table~\ref{tab:robustP8FixSet} and Table~\ref{tab:robustP11FixSet}.
We observe similar results as the robustness verification under Hamming distance.

\begin{table}[t] \setlength{\tabcolsep}{7pt}\renewcommand\arraystretch{1.1}
	\centering
	\caption{Class distribution of samples on P8 under the set of indices $I_8$}\label{tab:robustP8FixSet}
	\begin{adjustbox}{width=.7\textwidth,center}
	\begin{tabular}{   c | rrrrr rrrrr }
		\toprule
		Region & \ 0 \ & 1 & 2 & 3 & 4 & 5 & 6 & 7 & 8 & 9 \\
		\midrule  \rowcolor{gray!20}
	   0 \ & 977 & 0 & 0 & 25 & 0 & 22 & 0 & 0 & 0 & 0 \\
		 1 \ & 2 & 0 & 90 & 818 & 0 & 35 & 18 & 0 & 61 & 0  \\ \rowcolor{gray!20}
		 2 \ & 0 & 0 & 582 & 367 & 0 & 0 & 0 & 0 & 75 & 0 \\
		 3 \ & 30 & 0 & 33 & 883 & 0 & 50 & 0 & 0 & 28 & 0  \\	 \rowcolor{gray!20}	
		 4 \ & 0 & 8 & 128 & 306 & 200 & 98 & 176 & 17 & 27 & 64 \\
		 5 \ & 0 & 0 & 0 & 865 & 0 & 72 & 0 & 3 & 84 & 0\\ \rowcolor{gray!20}
		 6 \ & 13 & 0 & 1 & 14 & 0 & 17 & 967 & 0 & 2 & 10 \\
		 7 \ & 1 & 0 & 0 & 60 & 0 & 40 & 0 & 695 & 89 & 139  \\ \rowcolor{gray!20}
		 8 \ & 0 & 0 & 1 & 2 & 0 & 460 & 1 & 3 & 552 & 5  \\
		 9 \ & 26 & 0 & 0 & 6 & 11 & 41 & 111 & 65 & 172 & 592 \\
		\bottomrule
	\end{tabular}
	\end{adjustbox}
\end{table}

\begin{table}[t] \setlength{\tabcolsep}{0pt}\renewcommand\arraystretch{1.1}
	\centering
	\caption{Class distribution of samples on P11 under the set of indices $I_{11}$}\label{tab:robustP11FixSet}
	\begin{adjustbox}{width=\textwidth,center}
	\begin{tabular}{   c | cccc cccccc}
		\toprule
		Region & \ 0 \ & 1 & 2 & 3 & 4 & 5 & 6 & 7 & 8 & 9 \\
		\midrule  \rowcolor{gray!20}
		 0 \ & 268,400,472 & 0 & 16 & 17 & 0 & 0 & 34,948 & 3 & 0 & 0 \\
		1 \ & 0 & 203,530,394 & 64,879,960 & 25,046 & 0 & 9 & 0 & 47 & 0 & 0 \\  \rowcolor{gray!20}
		2 \ & 0 & 0 & 268,377,162 & 3,138 & 0 & 0 & 0 & 8,875 & 46,281 & 0 \\
		3 \ & 0 & 0 & 13,123 & 268,422,333 & 0 & 0 & 0 & 0 & 0 & 0 \\ \rowcolor{gray!20}
		4 \ & 0 & 4,120 & 2,357,278 & 582,468 & 248,819,744 & 332,024 & 0 & 14,152,761 & 2,187,061 & 0  \\
		5 \ & 0 & 61,900 & 29,443 & 158,627,680 & 0 & 109,715,682 & 0 & 751 &0  & 0 \\ \rowcolor{gray!20}
		6 \ & 0 & 1,518 & 10,753 & 0 & 641 & 0 & 268,422,544 & 0 & 0 & 0 \\
		7 \ & 0 & 0 & 0 & 0 & 0 & 0 & 0 & 268,435,346 & 110 & 0 \\ \rowcolor{gray!20}
		8 \ & 0 & 974,769 & 344,245 & 14,759,485 & 0 & 0 & 0 & 0 & 252,356,957 & 0\\
		9 \ & 0 & 14 & 0 & 4 & 855 & 166 & 0 & 1,120,825 & 198,841 & 267,114,751 \\
		\hline
	\end{tabular}
	\end{adjustbox}
\end{table}

\end{document}